\documentclass[NewProceedings,letterpaper]{ascelike-new}
\usepackage{float}
\usepackage[utf8]{inputenc}
\usepackage[T1]{fontenc}
\usepackage{lmodern}
\usepackage{graphicx}
\usepackage{ragged2e}
\usepackage[figurename=Fig.,labelfont=bf,labelsep=period]{caption}
\usepackage{subcaption}
\usepackage{amsmath}
\usepackage{newtxtext,newtxmath}
\usepackage{caption}
\usepackage[colorlinks=true,citecolor =red,linkcolor=black]{hyperref}
\begin{document}

% You will need to make the title all-caps
\title{Towards an AI-based Early Warning System for Bridge Scour}

\author[1]{Negin Yousefpour}
\author[2]{Oscar Correa}

\affil[1]{Department of Infrastructure Engineering, The University of Melbourne. Email: negin.yousefpour@unimelb.edu.au}
\affil[2]{School of Computing and Information Systems, The University of Melbourne. Email: oscar.correa@unimelb.edu.au}

\maketitle

% Please include an abstract:
\begin{abstract}
Scour is the number one cause of bridge failure in many parts of the world. Considering the lack of reliability in existing empirical equations for scour depth estimation and the complexity and uncertainty of scour as a physical phenomenon, it is essential to develop more reliable solutions for scour risk assessment. This study introduces a novel AI approach for early forecast of scour based on real-time monitoring data obtained from sonar and stage sensors installed at bridge piers. Long-short Term Memory networks (LSTMs), a prominent Deep Learning algorithm successfully used for time-series forecasting in other fields, were developed and trained using river stage and bed elevation readings for more than 11 years, obtained from Alaska scour monitoring program. The capability of the AI models in scour prediction is shown for three case-study bridges. Results show that LSTMs can capture the temporal and seasonal patterns of both flow and river bed variations around bridge piers, through cycles of scour and filling and can provide reasonable predictions of upcoming scour depth as early as seven days in advance. It is expected that the proposed solution can be implemented by transportation authorities for development of emerging AI-based early warning systems, enabling superior bridge scour management.

\end{abstract}

\section{Introduction}
Scour at bridge piers has been a global challenge. It is the number one cause of bridge failure in many countries. With more than 26,000 scour critical bridges in the US, scour accounts for around 60\% of bridge failures~\cite{Wardhana:2003}. In Australia, many bridge failures have been reported in recent years, mainly due to extreme weather conditions, in particular flooding~\cite{Lebbe:2014}. The annual probability of a scour-related bridge failure in the UK has been estimated at around 30\%~\cite{van2014flood,Dikanski2016}. In Taiwan, in 2009 alone, approximately 150 bridges collapsed during extreme flooding~\cite{Hongetal.:2012}.

% An investigation by the US Federal Highway Administration (FHWA) in 2015 reported that $~$\$16.4 out of the total ~\$17.5 billion budget allocated to bridge construction was spent on rehabilitation or replacement of existing bridges \cite{FHWA:2016}. To reduce the cost of bridge maintenance and reduce failure risks, implementation of health monitoring and early warning systems has been accelerated in the past decade. In 2006, FHWA initiated a major program called Long-Term Bridge Performance with the objective of instrumenting and monitoring a large number of bridges over a 20-year period \cite{Friedland:2019,Kirk:2018,Parvardeh:2016}.%
 
Scour is known as a highly uncertain physical phenomenon due to complexity of flow, structure, and soil interaction. Despite numerous efforts, the assessment of existing empirical scour models shows that the majority can result in significant errors, in particular over-prediction of scour. These regression models have been calibrated based on limited laboratory test results under certain conditions with limited field verification~\cite{FHWA:2005,Sheppard:2014}. The most commonly used scour equation for bridge design has been the HEC-18, which has evolved throughout the past two decades~\cite{Hec-18:2012}. A recent review of the HEC-18 model by  Schuring et al.~\citeyear{Hec-18Rev:2017} reveals several limitations, the most important of which are (I) insensitivity to a broad range of geological conditions, especially very coarse granular particles, (II) impracticality of proposed solutions for cohesive soils, (III) lack of recommendations for future scour potential for existing bridges, i.e. field performance over the time.

To date, numerous research studies have worked on development of scour predictive models using "classical" Machine Learning (ML), as opposed to "modern" ML or Deep Learning (DL), trained based on both lab and field data~\cite{Zounemat:2009,Azamathullaetal:2010,Kimetal:2015,Ebtehajetal:2016}. Sharafati et al.~\citeyear{sharafati:2021} have provided a comprehensive review on the most recent ML models for scour and their merits and limitations. These models generally provide more accurate maximum scour depth predictions compared with the empirical scour models. However, like any classical ML model, their extrapolation capacity is limited and the accuracy of predictions decay significantly outside the \textit{convex hull} of the training dataset, i.e. poor generalization to unseen data outside the training data~\cite{McCartneyetal.:2020,Yousefzadeh:2022}. Moreover, these models are often high-dimensional, incorporating a large number of bridge, flow, and soil characteristics as the predictive variables (input parameters), and are often trained using data compiled from various scattered locations, therefore only validated under certain data domain ranges. These factors further expose these models to extrapolation risk, rendering less reliability in scour predictions for an unseen bridge~\cite{Balestriero:2021}. 

In addition to limitations of the current scour prediction models, there are other factors to blame for scour failure, among which are: lack of proper design and analysis method for older bridges subjected to scour; hydraulic uncertainties, especially increase in flood intensity and climate change impacts~\cite{khandel:2019}; and complexity of riverbed sediment and morphological behaviour~\cite{lamb:2017}. Given the aforementioned limitations, bridge authorities have resorted to monitoring solutions to manage the scour risk. This includes both regular surveys and inspections as well as real-time monitoring using advanced sensor technologies.

Real-time scour monitoring have become increasingly prevalent in recent years and have enabled more reliable scour risk management for critical and large-scale bridges~\cite{NCHRP:1997,NCHRP:2009,Briaud:2011,PRENDERGAST:2014}. Chang et al.~\citeyear{CHANG:2014} proposed a real-time monitoring solution using micro-cameras installed at bridge piers and advanced image processing and pattern recognition methods to monitor bed level variations in real-time. They validated the method through laboratory experiments and reported that image data reliably captured monitored scour-depth evolution. Maroni et al.~\citeyear{Maroni:2020} proposed a scour hazard model for road and railway bridges, using Bayesian networks calibrated based on limited data from scour monitoring systems on a bridge network in Scotland. They reported the Bayesian network approach to significantly reduce the uncertainty in the scour depth assessment at unmonitored bridges. Lin et al.~\citeyear{Linetal.:2021} introduced a new scour monitoring technology using vibration-based arrayed sensors, Internet of Things (IoT), and artificial intelligence (AI). The water level changes around the bridge pier were captured using real-time CCTV images and processed using deep learning methods. The monitoring system was validated through laboratory flume experimentation and field measurements. 

Yousefpour et al.~\citeyear{Yousefpour2020scour} investigated scour prediction using AI/ML, based on monitoring data in collaboration with USGS and some of the US departments of transportation (DOTs). Three ML methods for real-time forecast and prediction of maximum depth of scour were developed. The developed models showed promising results in scour depth prediction based on real-time monitoring data. The data included river bed elevation measurements through \textit{sonar} sensors ($y_{so}$) and river stage (flood/water level) measurements through \textit{stage} sensors ($y_{st}$) installed on more than 20 bridges in Alaska between 2001 to 2017. The preliminary results showed great potential for Deep Learning (DL) using Long-short term memory networks (LSTMs) for real-time scour forecast. 

Following this study, in this paper we introduce an innovative AI-based, early warning solution for scour assessment that can be integrated with the real-time scour monitoring systems, such as those installed for many bridges in various states in the USA (e.g. Alaska, Oregon, Colorado, Idaho). Novel LSTM algorithms are developed and trained to recognize the inherent scour and flood patterns in the historic monitoring data, i.e. time series of river stage and bed elevation. The AI models can process the monitoring data in real-time and generate intelligent forecasts on upcoming scour depth enabling a superior scour risk management system. 

The essence of our research is creating a bridge-specific expert system for scour assessment, based on its historical and real-time sonar and stage sensor readings without the need to involve all the physical parameters typically used in  scour equations, such as pier geometry, velocity and soil condition (particle size). We base our data-driven solution on the fact that DL models can learn complex physical processes from the data, without exposure to explicit physics knowledge of the process~\cite{Kiarash:2019,Schwartz:2021}, while proving our hypothesis which is: \textit{the historic time-series of scour and flow depth and their cross-correlations embody the physical interactions of the soil, flow, and structure for a given bridge, which can be assimilated by modern deep-learning methods to provide useful insights on upcoming scour events}. 

The theoretical bases for such hypothesis and the underlying assumptions lies with the scour physical process: (I) the DL models are developed to be bridge-specific, i.e. learning cannot be transferred to another bridge, therefore the physical parameters that are site-specific and not temporally variable, such as bridge/pier geometry and soil condition/geology are not incorporated; (II) the flow velocity is directly correlated with the flow depth ($\approx y_{st}-y_{so}$), therefore it is not considered as an input to the AI model (see velocity impact analysis in section~\ref{sec: velocity}); (III) the soil condition impacts on the scour process is implicitly encoded in the historic river bed and flow variation patterns. 

%In this study we employ the unique capability of LSTMs in learning the inherent physics of the scour process within the historic time-series of sonar and stage sensor data for a specific bridge. 
Note that the aim of this research is not generating a ML model for maximum scour depth prediction, rather introducing an early-warning solution by integrating AI with the scour monitoring technology to provide predictive insights for risk assessments of upcoming scour events. The proposed solution is demonstrated in the following sections for three case study bridges in Alaska: Chilkat River (bridge ID: 742), Knik River (bridge ID: 539), and Sheridan Glacier No.3 (bridge ID: 230). These were the bridges with the most significant scour and filling associated with seasonal high flows and flood events, and for which the sensor readings were consistently available.

\section{Approach}
\subsection{Scour Monitoring Data}
\label{sec: datasets}

The sonar and stage sensor readings have been provided by USGS and DOTs accompanied by short cover reports highlighting the major scour and flooding events, newly installed sensors, and replacements or damages to sensors on monitored bridges, among other details. The real-time monitoring data for Alaska bridges can be accessed through the online monitoring platform available at \href{https://www.usgs.gov/centers/asc/science/streambed-scour-bridges-alaska?qt-science_center_objects=0#qt-science_center_objects}{USGS Alaska Science Center}. 
 
Prior to using the sensor data for model development, extensive pre-processing was performed to generate smooth continuous time series with synchronized steps between sonar and stage readings.  In some cases, a small portion of data was missing due to issues with the sensors (damages due to debris, vandalism, etc.) or the readings were noisy and erroneous. Also, sensor reading intervals were not uniform across all the bridges and years of monitoring. The following main steps were taken for pre-processing of data. See \cite{Yousefpour2020scour} for a full description of the data and pre-processing methods:

\begin{itemize}
\item	Apparent bias shifts in the sensors were corrected based on manual inspection.
\item	Outliers were removed via median filtering.
\item	Timestamps were made uniform (1hr) throughout the dataset and synchronized between sonar and stage.
\item	Denoising/filtering technics including Moving Average and Low Pass Filter were applied to remove high-frequency noises and to capture the actual trend. 
\item	Missing data were imputed using polynomial interpolation and Gaussian Process.
\end{itemize}

\begin{figure}[H]
        \centering
        \includegraphics[width=\textwidth]{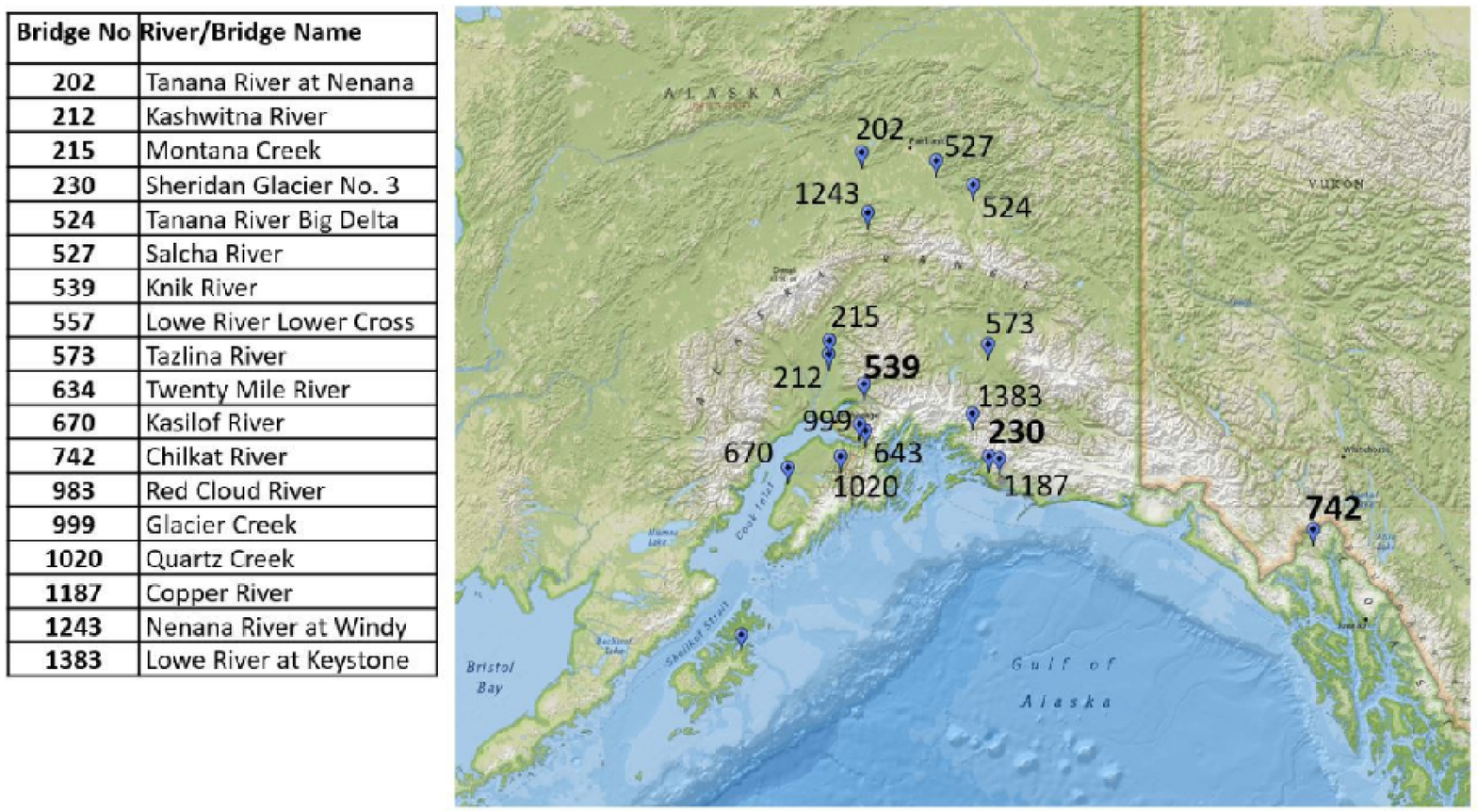}
        \caption{Map of Alaska bridges with active real-time monitoring systems, case study bridges highlighted}
        %(from: \href{https://www.usgs.gov/centers/asc/science/streambed-scour-bridges-alaska?qt-science_center_objects=0#qt-science_center_objects}{USGS Alaska Science Center})~\cite{USGS2022}}
        \label{fig:map}
\end{figure}

Figure \ref{fig:map} shows the map of the bridges with installed sensors across Alaska as part of the USGS real-time monitoring program. Sounding survey results from 2001 to present with an image of the bridge with installed sensors for bridges 742, 539 and 230, are shown in Figure \ref{742-sounding}, Figure \ref{539-sounding}, and Figure \ref{230-sounding} respectively. Also, Figure \ref{742-sample data}, Figure \ref{539-sample data}, and Figure \ref{230-sample data} provide the raw versus processed data for recent years. Note that all the original data were provided in US units (ft for sensor readings of elevation). Exploring the data shows similar yearly periodicity across bridges in Alaska. Most bridges experience two major high flow/flooding around July/August and September/October. There is no reading due to frozen waters between November and March, hence the gaps (or straight lines) between the years. 

\subsubsection{Bridge 742 - Chilkat River}
\begin{figure}[H]
   \centering
      \begin{subfigure}{0.7\textwidth}
      \centering
      \includegraphics[width=\textwidth]{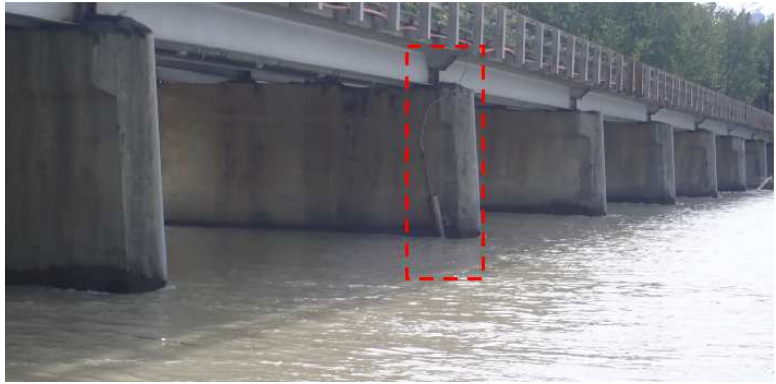}
      \caption{}
      \end{subfigure}
      \begin{subfigure}{\textwidth}
      \centering
      \includegraphics[width=\textwidth]{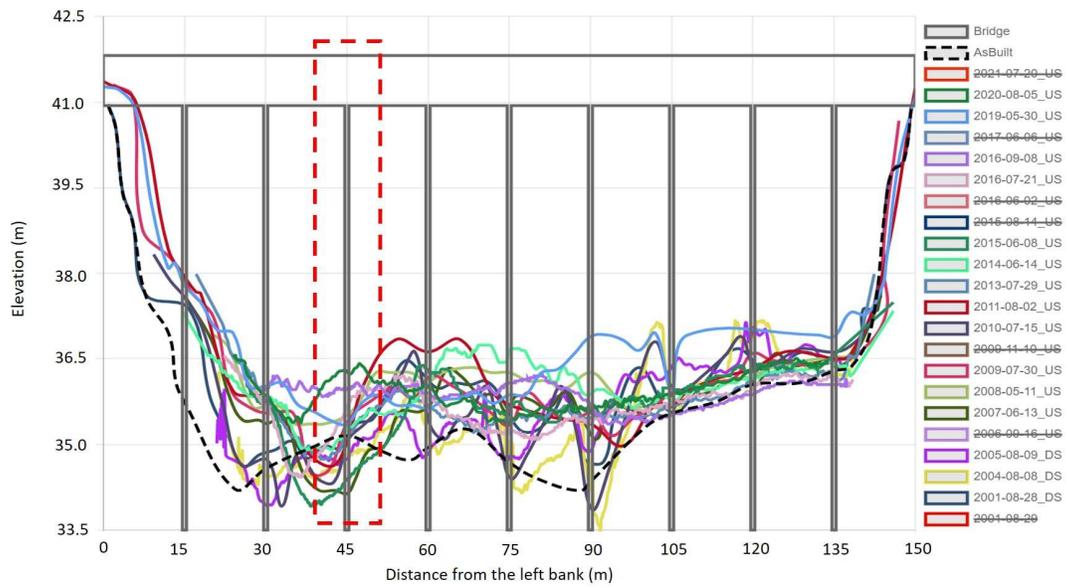}
      \caption{}
      \end{subfigure}
  \caption{Scour monitoring of bridge 742, a) a photo showing sensors mounted on the third pier from the left bank} 
  %(from: \href{https://www.usgs.gov/centers/asc/science/streambed-scour-bridges-alaska?qt-science_center_objects=0#qt-science_center_objects}{USGS Alaska Science Center}), b) sounding surveys from 2001 (original units converted from ft to m)~\cite{USGS2022}}
\label{742-sounding}
\end{figure}

\begin{figure}[H]
   \centering
      \begin{subfigure}{1\textwidth}
      \centering
      \includegraphics[width=\textwidth]{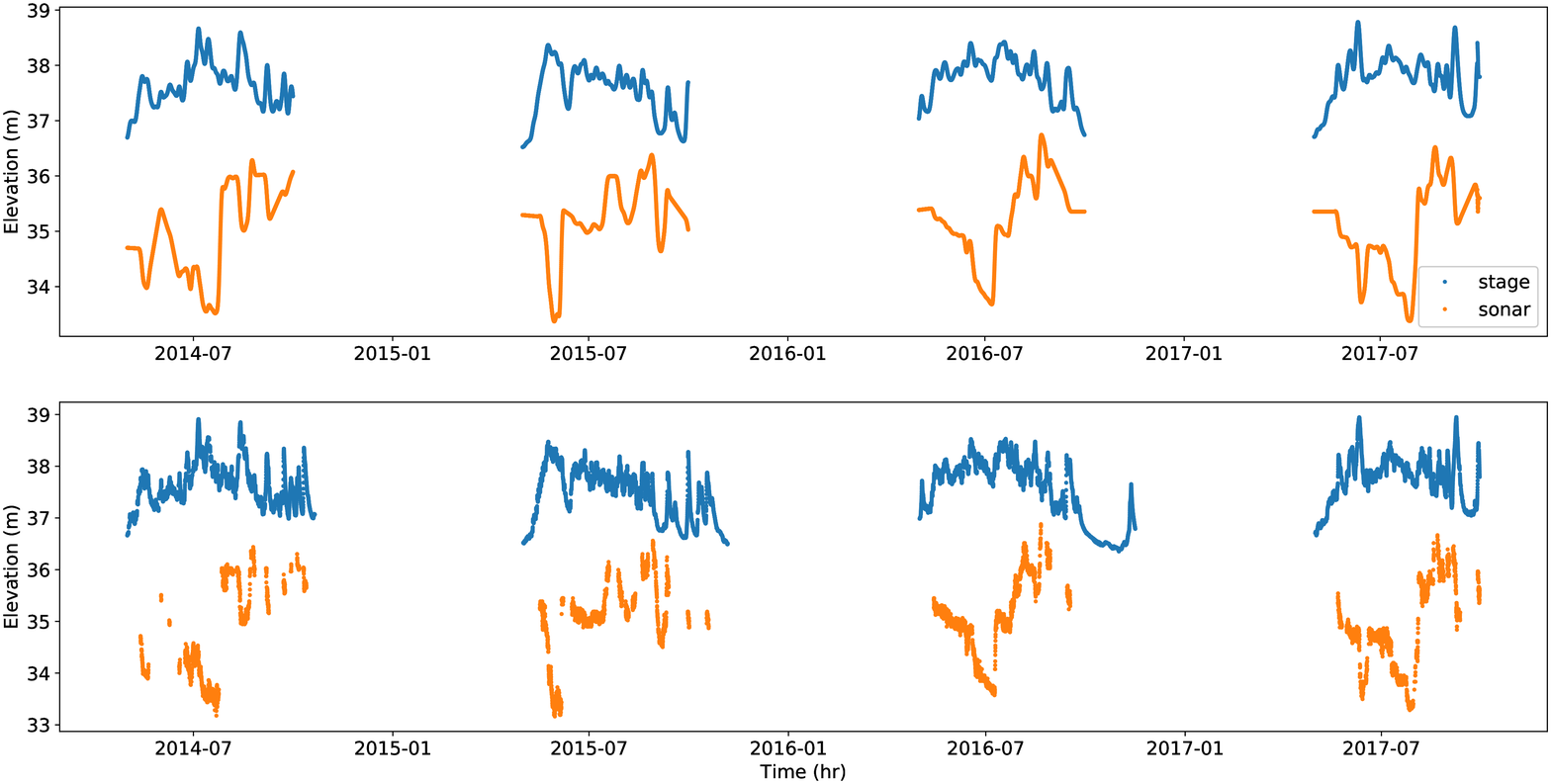}
      \caption{}
      \end{subfigure}
      \begin{subfigure}{1\textwidth}
      \centering
      \includegraphics[width=\textwidth]{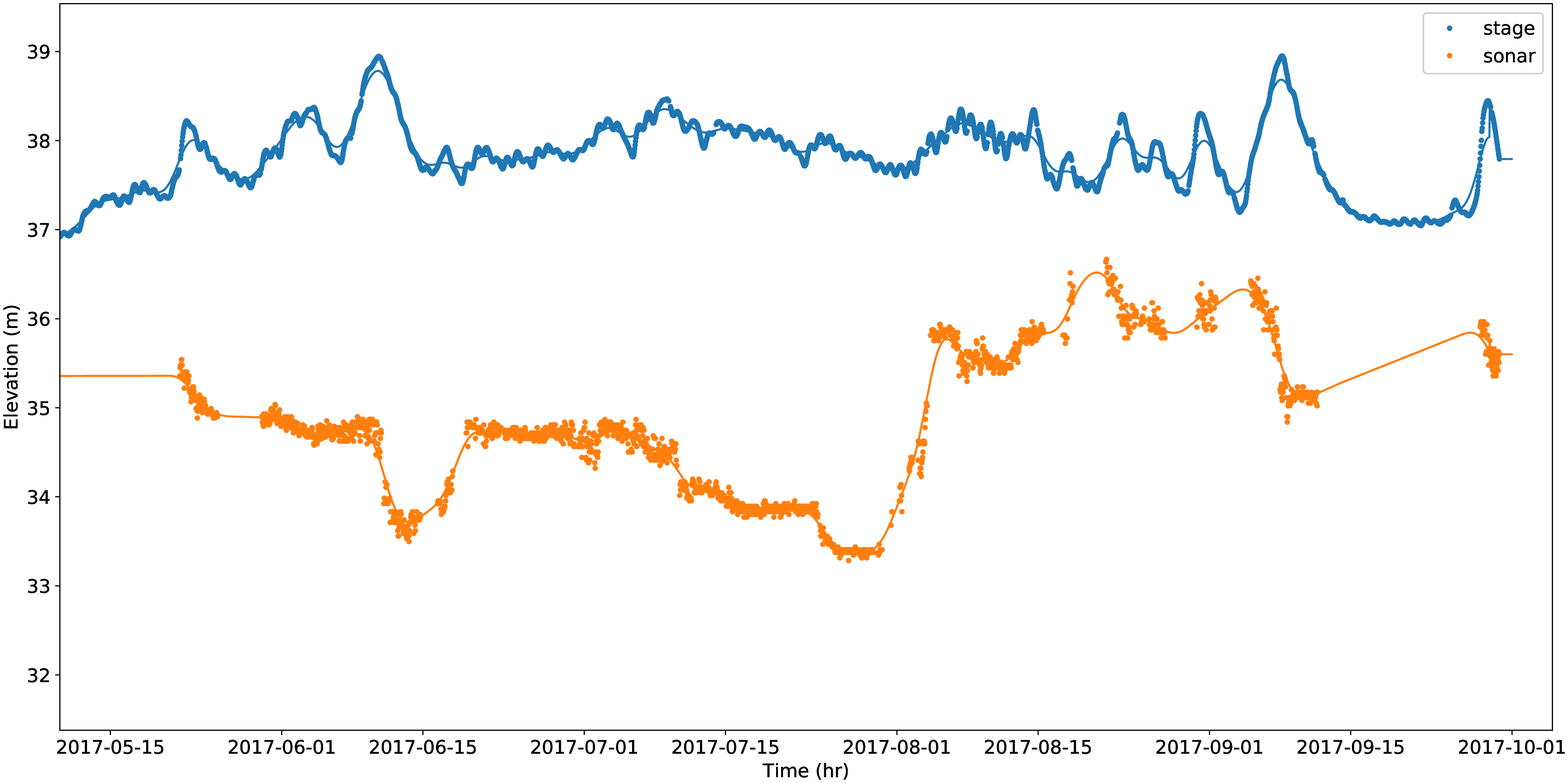}
      \caption{}
      \end{subfigure}
   \caption{Bridge 742 - sample of sensor data from continuous real-time monitoring: a) top: processed; bottom: raw, b) 2017 data - processed (line) and raw (dots)} 
\label{742-sample data}
\end{figure}

As observed from Figure \ref{742-sounding} and Figure \ref{742-sample data}, historical sounding and sensor data for bridge 742 shows cyclic scour and filling of up to %7-10ft
4m due to seasonal high flows (floods) happening around July/August and September/October each year and the bed elevation fluctuating between %110ft 
33m to %120ft 
37m (see also Figure \ref{fig: datadiv-742}). The sounding surveys allowed for cross-validating the readings from sonar and ensuring the maximum depth of scour is captured across the channel. Sonar has been installed on the third pier from the left bank, which is closest, among the nine piers, to the major scour hole across the river bed. 

\begin{figure}[H]
    \centering
    \begin{subfigure}{1\textwidth}
    \centering
    \includegraphics[width=0.5\textwidth]{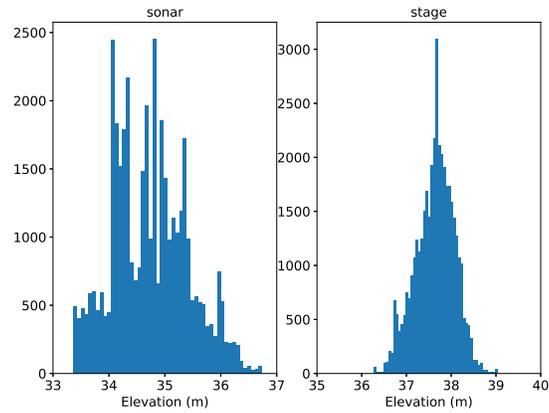}
    \caption{}
    \end{subfigure}
    \begin{subfigure}{1\textwidth}
    \centering
    \includegraphics[width=0.5\textwidth]{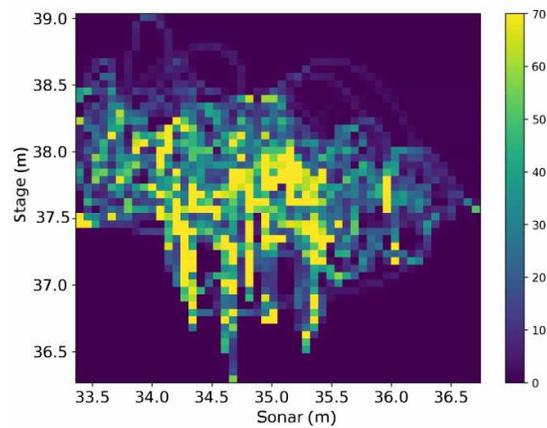}
    \caption{}
    \end{subfigure}
    \caption{a) Individual and b) joint histograms of sonar and stage readings - bridge 742}
    \label{fig:742-Hist}
\end{figure}

Figure \ref{fig:742-Hist} provides the histogram of the processed stage and sonar readings for bridge 742 from 2007 to 2017. The negative correlation between the two features can be observed in the joint histogram. 

\subsubsection{Bridge 539 - Knik River}
\begin{figure}[H]
   \centering
      \begin{subfigure}{0.7\textwidth}
      \centering
      \includegraphics[width=\textwidth]{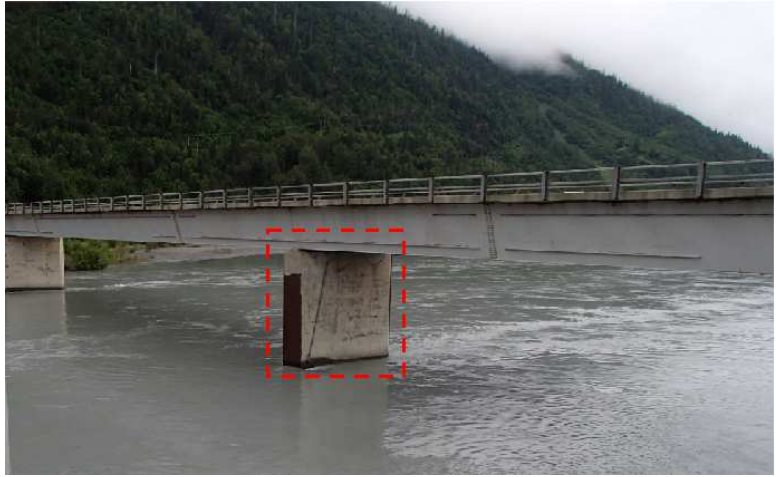}
      \caption{}
      \end{subfigure}
      \begin{subfigure}{1\textwidth}
      \centering
      \includegraphics[width=\textwidth]{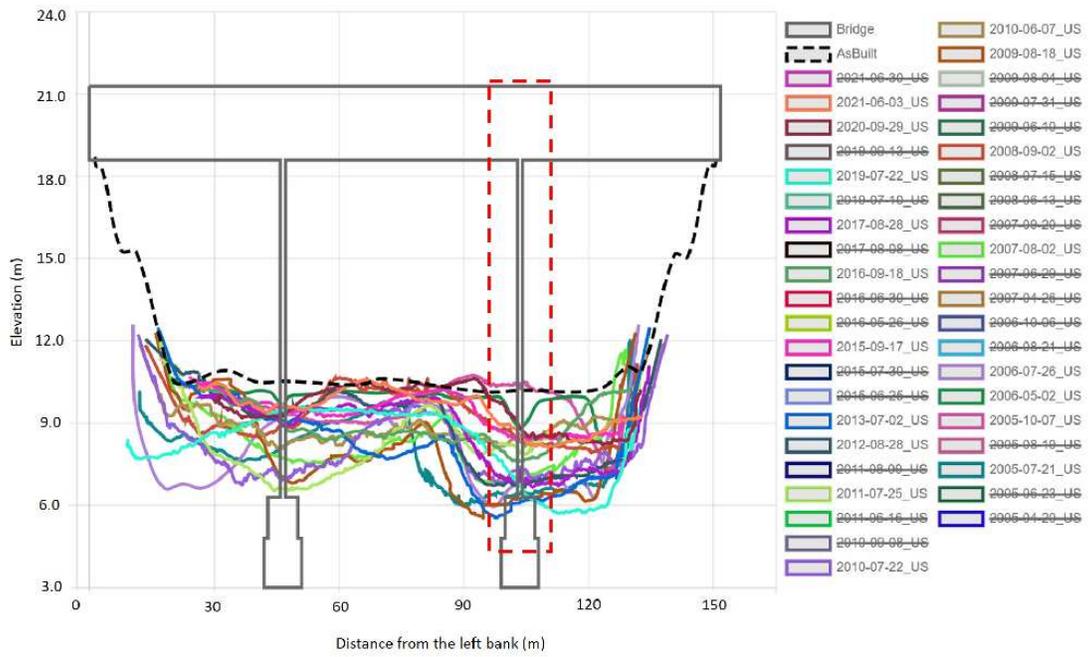}
      \caption{}
      \end{subfigure}
\caption{Scour monitoring of bridge 539, a) a photo showing sensors mounted on pier closest to the right bank 
%(from: \href{https://www.usgs.gov/centers/asc/science/streambed-scour-bridges-alaska?qt-science_center_objects=0#qt-science_center_objects}{USGS Alaska Science Center}), b) sounding surveys from 2005 (units converted from ft to m)~\cite{USGS2022}
}
\label{539-sounding}
\end{figure}

\begin{figure}[H]
   \centering
      \begin{subfigure}{1\textwidth}
      \centering
      \includegraphics[width=\textwidth]{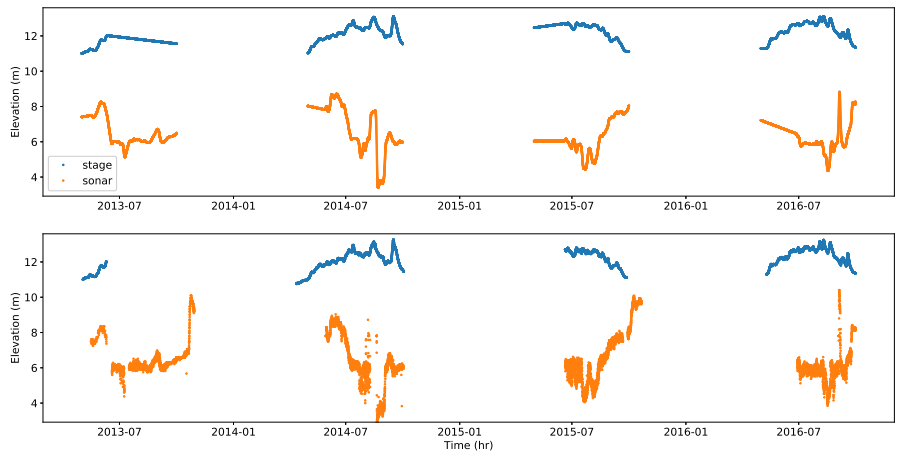}
      \caption{}
      \end{subfigure}
      \begin{subfigure}{1\textwidth}
      \centering
      \includegraphics[width=\textwidth]{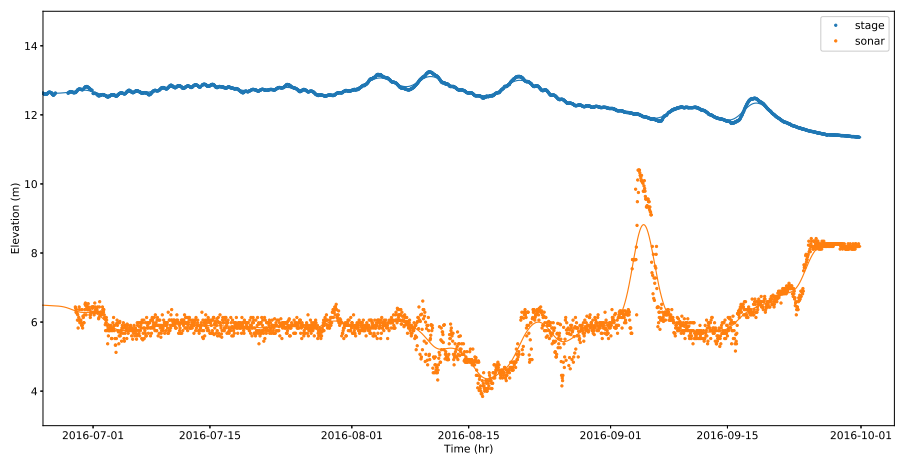}
      \caption{}
      \end{subfigure}
   \caption{Bridge 539 - Sample of sensor data from continuous real-time monitoring, a) top: processed; bottom: raw, b) 2016 data - processed (line) and raw (dots)} 
\label{539-sample data}
\end{figure}

As shown in Figure \ref{539-sounding} and \ref{539-sample data}, bridge 539 goes through cyclic scour and filling between summer and fall up to a maximum of %10-15ft 
6m, with the river bed fluctuating between %10ft 
4m to %35ft 
10m elevations (see also Figure \ref{fig: datadiv-539}). The sonar has been installed on the pier closest to the right bank. 

Due to the 2006 major flood and damage to sensors, the Stage readings were unavailable in 2006.

\subsubsection{Bridge 230 - Sheridan Glacier}
\begin{figure}[H]
   \centering
      \begin{subfigure}{0.7\textwidth}
      \centering
      \includegraphics[width=\textwidth]{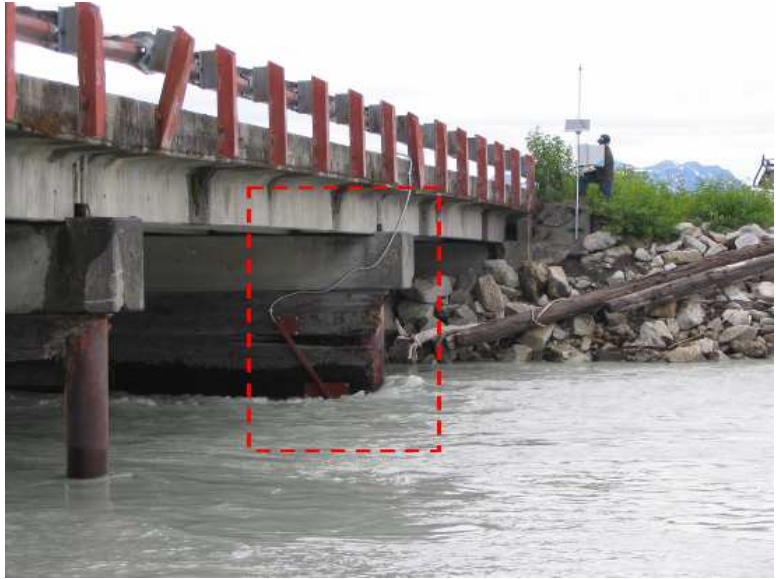}
      \caption{}
      \end{subfigure}
      \begin{subfigure}{1\textwidth}
      \centering
      \includegraphics[width=\textwidth]{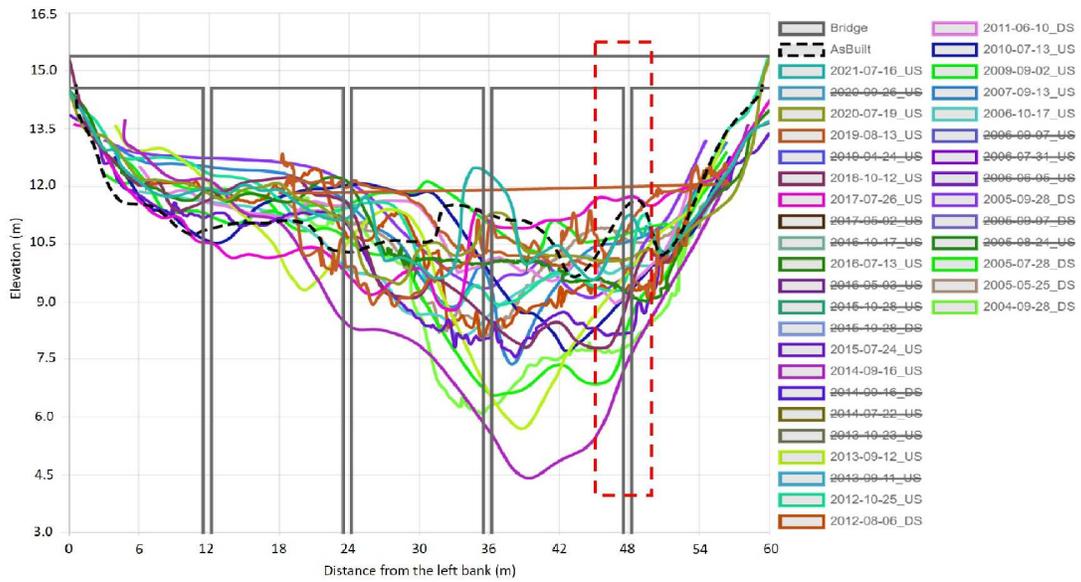}
      \caption{}
      \end{subfigure}
 \caption{Scour monitoring of bridge 230, a) a photo showing sensors mounted on pier closest to the right bank 
 %(from: \href{https://www.usgs.gov/centers/asc/science/streambed-scour-bridges-alaska?qt-science_center_objects=0#qt-science_center_objects}{USGS Alaska Science Center}), b) sounding surveys from 2004 (original units converted from ft to m)~\cite{USGS2022}
 }
\label{230-sounding}
\end{figure}

\begin{figure}[H]
   \centering
      \begin{subfigure}{1\textwidth}
      \centering
      \includegraphics[width=\textwidth]{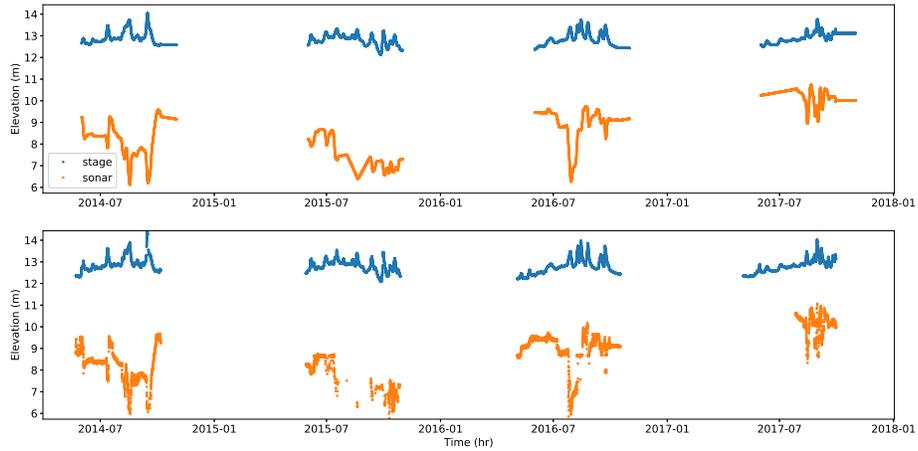}
      \caption{}
      \end{subfigure}
      \begin{subfigure}{1\textwidth}
      \centering
      \includegraphics[width=\textwidth]{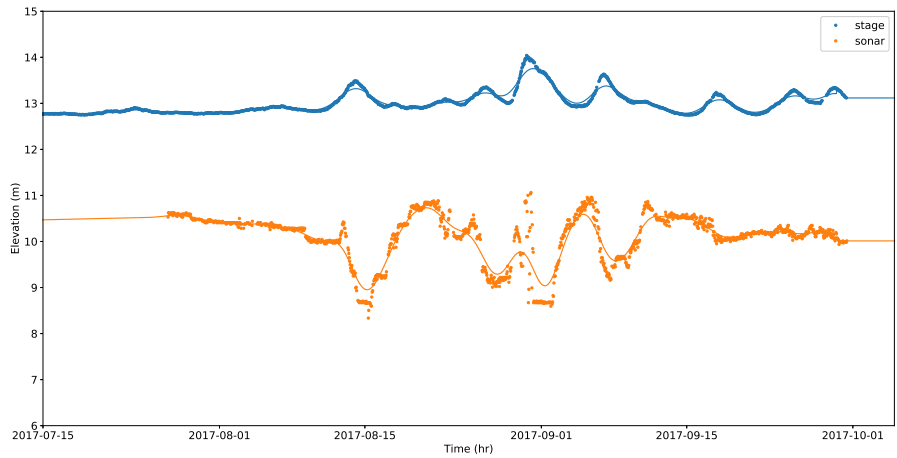}
      \caption{}
      \end{subfigure}
   \caption{Bridge 230 - sample of sensor data from continuous real-time monitoring, a) top: processed; bottom: raw, b) 2017 data - processed (line) and raw (dots)}  
\label{230-sample data}
\end{figure}

Bridge 230 has gone through several over-topping episodes during major floods between 2001 to 2017, with a flood exceeding 100-year return period in 2006. 
As shown in Figure \ref{230-sounding} and Figure \ref{230-sample data}, the sonar has been installed on the pier closest to the right. The sonar recordings show up to 5m cycles of scour and filling and bed elevation variation between 6m to 11m (see also Figure \ref{fig: datadiv-230}). 

Due to the sensors malfunction, the readings were not used in 2009 and 2010.

\subsection{Early Warning Solution Framework}
\begin{figure}
    \centering
    \includegraphics[width=\textwidth]{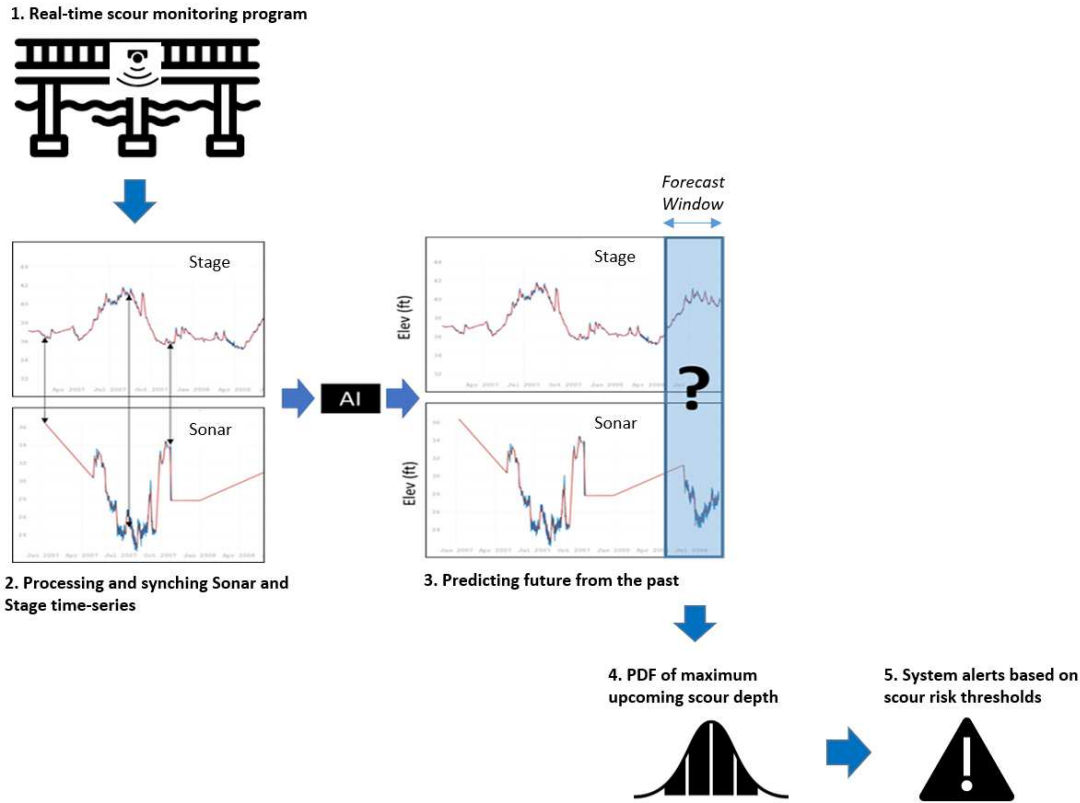}
    \caption{AI-Based early warning framework for bridge scour using real-time monitoring data}
    \label{fig:EarlyWarning}
\end{figure}

Figure \ref{fig:EarlyWarning} presents a schematic of the AI early warning solution framework we propose in this paper. The first and second steps i.e. real-time monitoring of scour and pre-processing of the historic scour monitoring data were explained in the previous section. The third step is developing AI algorithms that learn from past to predict the future, i.e. the variation of bed elevation across a pre-defined forecast window. In the forth step, the upcoming maximum scour depth is estimated based on the probability distribution function (PDF) of predictions generated by the AI model. Finally, in the last step the predicted scour depth is evaluated with respect to a set of threshold defined based on failure risk assessments, considering the as-built design of the foundation. This process involves subtracting the predicted scour depth for a target probability of exceedance (e.g. $<$ $10\%$) from the as-built embedment depth of the foundation to assess the residual bearing capacity. In this paper we focus on step-3, where we introduce LSTM networks as surrogate models for scour forecast as explained in the following sections.

\subsection{Deep Learning}

One of the most prominent deep learning models to approach a time series forecasting problem is Recurrent Neural Networks (RNNs), capable of "remembering". In addition to feed-forward connections between computational units (neurons), RNNs have recurrent connections where the output of a unit is fed back to itself with a weight and a time delay, which provides the algorithm with a memory of past activations. This is the backbone idea of the solution proposed in this paper.

In processing the time-series data, RNNs iterate through the sequential elements (time-steps) while maintaining a \emph{memory state} of the activation at each step. The sequence, in this study, corresponds to a vector of two features: river stage (\texttt{stage}) and river bed elevation (\texttt{sonar}).
 
\subsubsection{Long Short-Term Memory Networks}

LSTM networks are special types of RNNs with an internal memory unit that can maintain short and long temporal patterns of past time-series data and incorporate them into future predictions ~\cite{hochreiter1997long,deng2014deep}. Stacking memory units in such networks enables learning higher levels of temporal patterns in sequential data. LSTMs have been successfully applied in a wide range of applications, such as stock market forecast, text, language, and voice recognition ~\cite{Rayetal.:2015}, and recently used in some geotechnical engineering applications ~\cite{Xie:2019,Zhangetal:2020,ZHANGetal:2021,Yousefpour:2022}. 

One of the main advantages of LSTMs over simple RNNs is that the LSTM memory unit can overcome the \emph{vanishing gradient} problem, as it allows the past information to be re-injected at a later time. Vanishing gradient is a common learning pitfall in deep recurrent neural networks, where the gradient of weights becomes too small or too large, if the network is unfolded for too many time steps \cite{bengio1994learning,hochreiter1997long}. 

\begin{figure}[H]
   \centering
    \begin{subfigure}{0.6\textwidth}
      \centering
      \includegraphics[width=\textwidth]{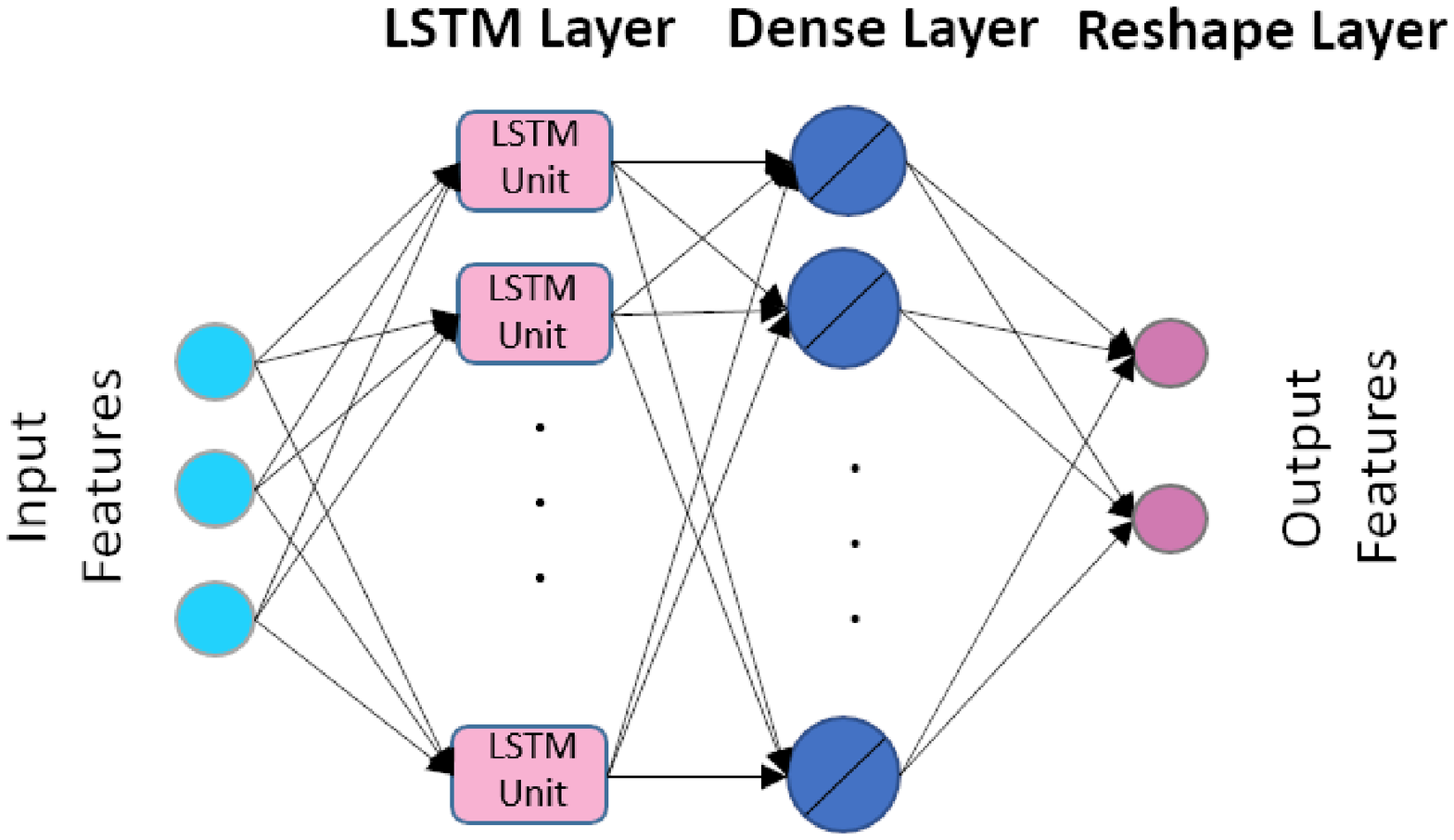}
      \caption{}
    \end{subfigure}
     \begin{subfigure}{0.6\textwidth}
      \centering
      \includegraphics[width=\textwidth]{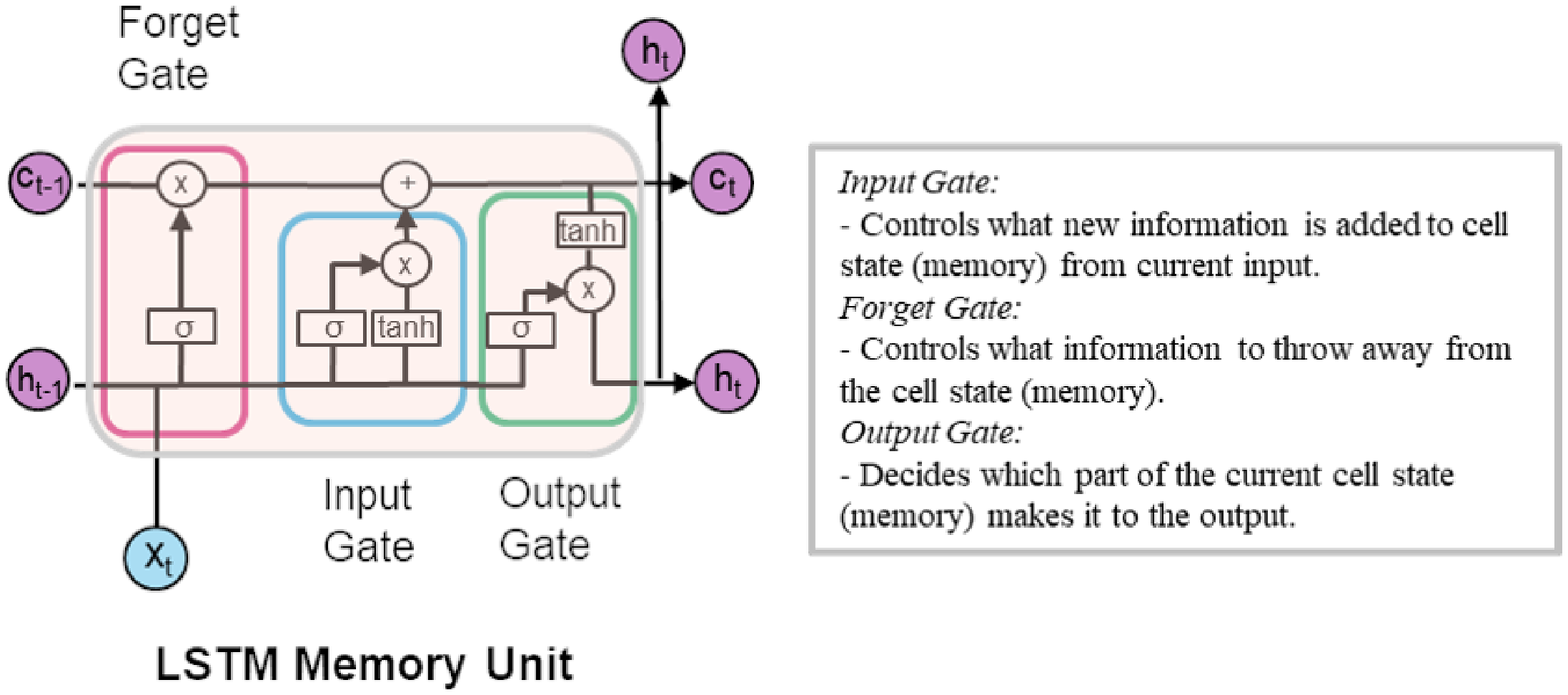}
      \caption{}
     \end{subfigure}
     \begin{subfigure}{0.7\textwidth}
      \centering
      \includegraphics[width=\textwidth]{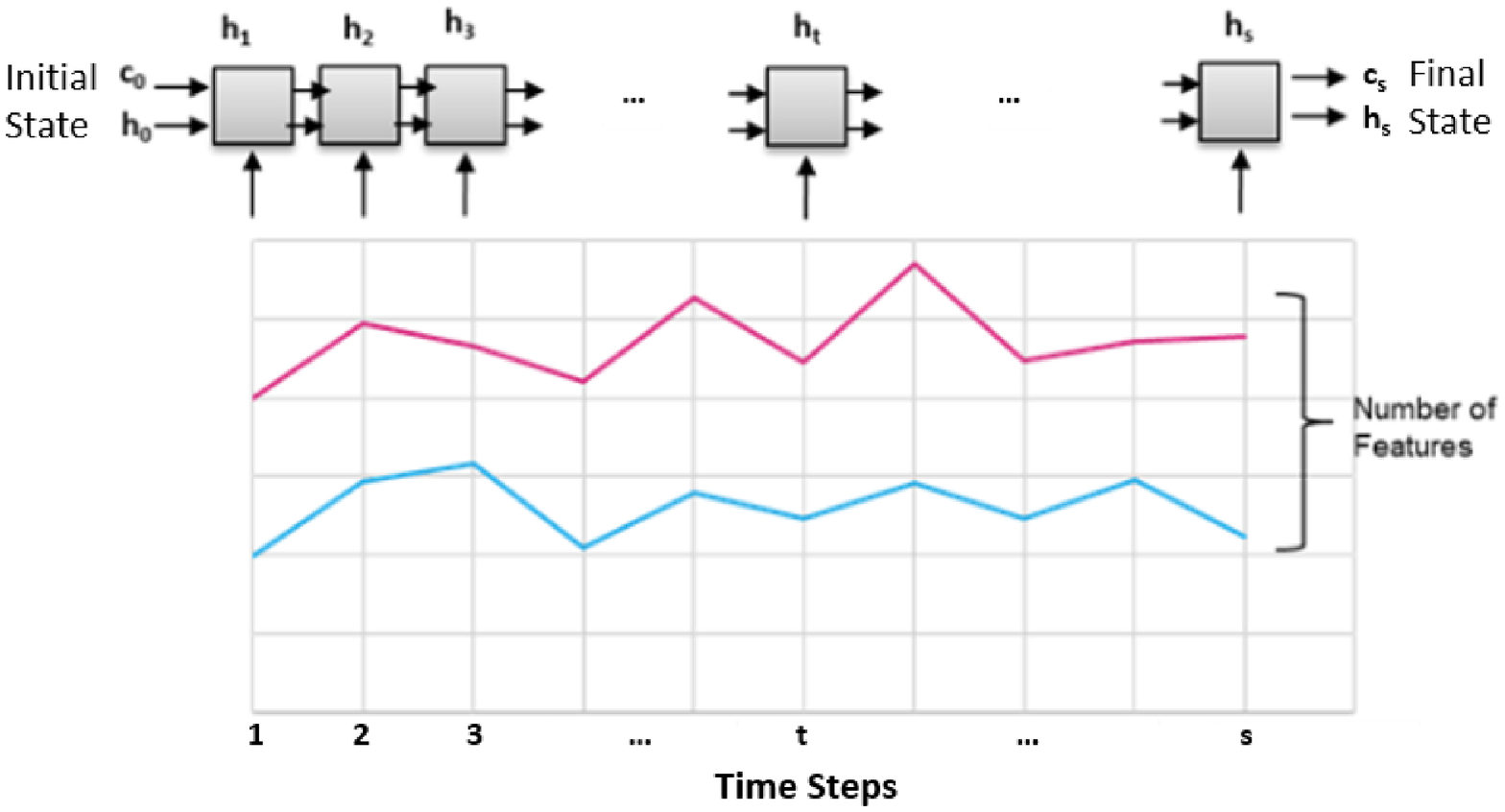}
     \caption{}
     \end{subfigure}
\caption{a) LSTM network architecture, b) LSTM unit architecture; symbols are as follows: $x_t$: input, $c_t$: cell memory state, $h_t$: hidden state or output at time step $t$; $\sigma$: sigmoid activation function; $tanh$: hyperbolic tangent function, c) LSTM unit unfolded in time.}
\label{LSTM-Arch}
\end{figure}

The LSTM algorithms in this study were developed using the TensorFlow library in Python. For time-intensive computations and configuration optimization, codes were run on several GPUs of Spartan, the high-performance computing system at The University of Melbourne. Figure~\ref{LSTM-Arch} shows a diagram of the LSTM architecture used in this study.

\begin{figure}
    \centering
     \begin{subfigure}{\textwidth}
     \centering
      \includegraphics[width=0.25\textwidth]{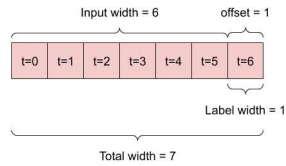}
      \caption{}
      \label{InputLabel}
     \end{subfigure}
     \begin{subfigure}{\textwidth}
     \centering
      \includegraphics[width=.25\textwidth]{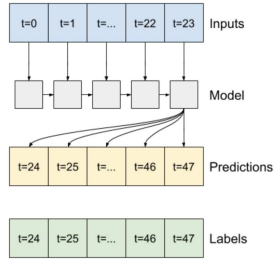}
      \caption{}
      \label{single-shot}
     \end{subfigure}
     \begin{subfigure}{\textwidth}
      \centering
      \includegraphics[width=.4\textwidth]{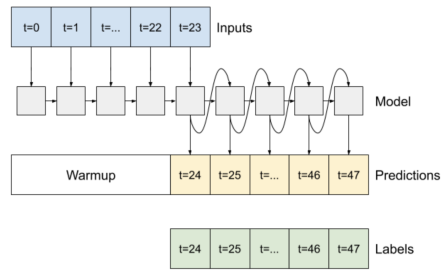}
      \caption{}
      \label{feedback}
     \end{subfigure}
  \caption{a) Definition of input, label and offset in time series data slicing scheme, b) Single-shot variant, c) Feedback \protect\cite {tftimeseries}}
  \label{fig: variants}
\end{figure}

To train the LSTMs, the time-series data are first sliced into smaller sequences through a sliding window. In each slice, a certain number of time-steps are treated as input (input width) to the network while a number of time-steps into the future is predicted (label width). The total length for each data sequence is the sum of the input and label widths (see Figure~\ref{InputLabel}). Note that no shift was considered in between the input and label chains in this study (offset width = label width). 

For the training process, the sequences in each subset of data (training, validation, and test) are divided into a number of batches. A maximum of 32 sequences are selected for each batch (batch size = 32). The LSTM performance is evaluated via the mean squared error (loss) and the mean absolute error (MAE) of predictions over the label width for all the batches in a dataset.  The network is trained using the training dataset over a large number of epochs (100) until the error (loss) on the validation dataset meets the stopping criteria. The most commonly activated stopping criteria is the threshold number for error increase recurrence (patience parameter). In addition to monitoring the training over the validation dataset, the \emph{Drop-out} technique is also explored to overcome the over-fitting issue. The performance of the models for unseen data is evaluated through the test dataset.

Three LSTM variants were investigated as explained in the following sections:

\begin{enumerate}
    \item Single-Shot (\texttt{ss})
    \item Feedback or Auto-regressive (\texttt{fd})
     \item Two-Layer, Single-Shot (\texttt{ss2})
\end{enumerate}

\subsubsection{Single-Shot Model}
In this variant, all the time steps are predicted at once with no offset as depicted in Figure \ref{single-shot}. The Python implementation of the model using TensorFlow comprises of three layers: an LSTM layer of units with sigmoid activation, a Dense layer, containing units with ReLU activation function: $max(0,x)$, and a Reshape layer, containing linear units to adjust the output dimensions at the end. The Dense layer has as many units as the number of time-steps to be predicted (\texttt{label\_width}) times the number of features (\texttt{n\_LabelFeat}).
The output from the Dense layer, which has two dimensions (\texttt{batch size, label\_width * n\_LabelFeat}), is reshaped into a tensor of three dimensions (\texttt{batch size, label\_width, n\_LabelFeat}), where the prediction of the label features is set up.

\subsubsection{Feedback (Auto-regressive) Model}
Similar to Single-Shot, this variant also has one LSTM layer, however, instead of predicting the output time-steps all at once, predictions are made one step at a time. The hypothesis is whether feeding back the model's output into itself at each step, i.e., conditioning prediction at each time-step on the previous one, increases the accuracy. As shown in Figure \ref{feedback}, the model predicts only one hour in the future, and then that prediction is the input for the prediction of the next time-step. 

An obvious advantage of this model is that the number of time-steps to be predicted in the future is unbounded.

\subsubsection{Two-Layer, Single-Shot Model}
This model is a Single-Shot variant with two LSTM layers. It is sometimes useful to stack several LSTM layers to increase the representational power of a network, as it is also common in feed-forward neural networks.
The trade-off is between the computational cost and representational power of the model.
Just to give an idea of how effective this strategy could be: Google Translate algorithms is a stack of seven large LSTM layers. It is generally a good idea to increase the network's capacity, i.e., stacking layers, until over-fitting becomes an obstacle.

\subsection{LSTM Configurations and Hyper-Parameters}
\label{sec: configuration}
A number of LSTM configurations were analysed in this study, considering possible combination scenarios for input features, model variants and other hyper-parameters. Using the \emph{grid-search} method, the most optimum models were identified by comparing numerous combinations of hyper-parameters. Table \ref{tab: hyperparams_values} presents the main hyper-parameters and their investigated values.

\begin{table}[H]
\caption{LSTM hyper-parameters and their values}
    \centering
    \begin{tabular}{c|c|c}
        \textbf{Hyper-Parameter} & \textbf{Range of Tested Values} &\textbf{Selected Value}\\
        \hline
        \hline
        Time Feature & Yes, No & No \\
        \hline
        No. of LSTM layers & 1, 2 & 1 \\ 
        \hline
		No. of hidden units & 32, 64, 128, 256 & Variable \\
	    \hline
	    Optimization Algorithm & SGDM, Adam and RMSProp & Adam \\
	    \hline
        Drop-out Rate & 0, 20\% & 0\\
        \hline
        Inputs, Labels (Sliding Window) & (336, 168), (720, 168), (720, 336) & (336, 168)\\
    \end{tabular}
    \label{tab: hyperparams_values}
\end{table}

\subsubsection{Features}
\label{sec: feature_engineering}
The scour data has clear yearly periodicity (e.g. frozen waters, seasonal variations of river stage/flooding and scour/filling), therefore we incorporate two additional features representing "time of the year" by applying $sin(\cdot)$ and $cos(\cdot)$ functions to time-stamps. Hence, the two feature combinations are:
\begin{enumerate}
    \item  \textit{With the time features} (\texttt{ssy}): \texttt{[sonar, stage, year sin, year cos]} 
    \item \textit{Without time features} (\texttt{ss}): \texttt{[sonar, stage]} 
\end{enumerate}

\subsubsection{Sliding Window}

The sliding-window hyper-parameter refers to the number of time-steps taken as input and label, i.e., input and label width. The following input-label widths were considered:

\begin{enumerate}
    \item (336, 168): Two weeks of history, one week of predictions
    \item (720, 168): Four weeks of history, one week of predictions
    \item (720, 336): Four weeks of history, two weeks of prediction
\end{enumerate}

Although initially increasing the input width may benefit the prediction accuracy, having long sequences will limit the number of sequences from the already small training dataset. A good trade-off between input width and the number of sequences is one of the key elements of optimal training. With respect to the number of steps to be predicted (label width), the problem itself sets a minimum range, which is 168 to 336 hrs (one to two weeks). This is considered as the minimum required time for bridge authorities to take countermeasures and devise a plan of action to manage the scour risk (confirmed in a meeting with various DOT bridge managers).

\subsubsection{Number of Epochs}

The number of epochs is set to 100, and the training was set to stop when monitored metrics stopped improving (the loss function stopped decreasing) . The parameter \texttt{patience} was set to 5, which is the number of consecutive epochs with no improvement, after which training is stopped.

\subsubsection{Number of Units}

Similar to feed-forward neural networks, increasing the number of units (hidden neurons) increases the predictive capacity of the network, yet at the risk of over-fitting and increased computational cost.
The chosen values for this parameter are powers of two: 32, 64, 128, and 256.

\subsubsection{Regularisation: Drop-out}

The drop-out rate is the proportion of units whose inputs are zeroed out in order to break random correlations in the training data. The drop-out version for RNNs is called \emph{recurrent drop-out}. Both drop-out and recurrent drop-out rates need to be optimally selected \cite{semeniuta2016recurrent}. Two values of 0 and 20\% were considered in the grid-search.

\subsection{LSTMs as Surrogate Models for Scour Forecast}
In this section we explain, through a mathematical representation, how the LSTMs can act as a surrogate model for physics-based, empirical equations for scour forecast. For this purpose we use HEC-18 as the most commonly used scour equation. Let's consider an individual pier as shown in Figure \ref{fig:Channel}, for an arbitrary bridge $N$. The scour depth (in SI units), $y_s$, can be written as function of two variables, $y_1$, the flow depth directly upstream of the pier and $V$, the mean velocity of flow directly upstream of the pier:
\begin{figure}[H]
    \centering
    \includegraphics[width=0.7\textwidth]{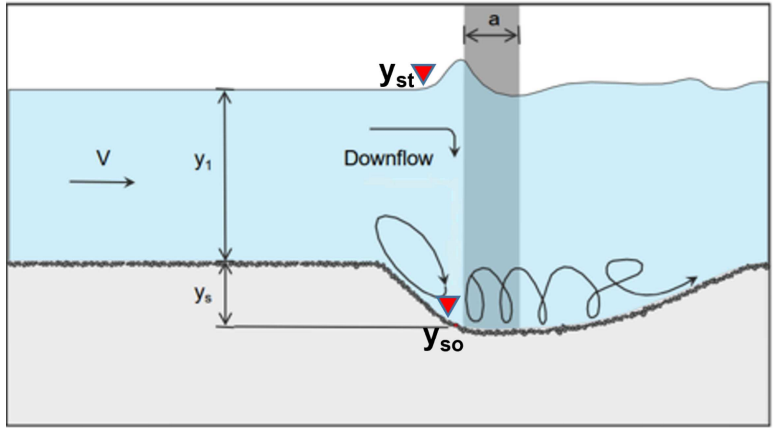}
    \caption{Schematic of flow and scour around a bridge pier, showing relevant parameters~\protect\cite{Hec-18:2012}}
    \label{fig:Channel}
\end{figure}

\begin{equation}
    \centering
    y_s=\alpha y_1^{0.13}V_1^{0.43}
    \label{Eq:Hec}
\end{equation}
where, $\alpha$ is a constant for bridge $N$, dependent on pier geometry, angle of attack, and river-bed condition at the pier location. Based on Manning’s equation for open channel flow, one can write:
\begin{eqnarray}
    \centering
    V&=&\frac{R^{0.67}s^{0.5}}{n}\\
    &=&\beta f(\mu y_1)
    \label{Eq:H=Mann}
\end{eqnarray}
where, $R$ is the hydraulic radius, which assuming the channel width to be constant, can be written as a function of flow depth: $f(\mu y_1)$, in which $\beta=s^{0.5}/n$, where $s$ and $n$ are the channel slope and Manning's roughness coefficient, and $\mu$ is a constant for bridge $N$ at the pier location.

Considering 
$\Big[\begin{smallmatrix}
y^i_{st} \\
y^i_{so} 
\end{smallmatrix}\Big]$,
the vector representing sonar and stage readings at time-step $t_i$, and $y'_{st}=\max\{y^i_{st}\}^k_{i=n}$ and $y'_{so}=\min\{y^i_{so}\}^k_{i=n}$, the maximum and minimum of readings corresponding to maximum flow and scour depth during the forecast window of $\textbf{t}=\{t^i\}^k_{i=n}=\{t^n,\cdots, t^k\}$, one can write:
\begin{equation}
 y_s+y_1 \approx y'_{st}-y'_{so}\longrightarrow y_1 \approx y'_{st}-y'_{so}-y_s
 \label{Eq:StSo}
\end{equation}
Replacing $y_1$ from Eq.\ref{Eq:StSo} in Eq.\ref{Eq:H=Mann}, and replacing $V$ from Eq.\ref{Eq:H=Mann} in Eq.\ref{Eq:Hec}, the maximum scour depth given by the empirical equation can be written as:

\begin{equation}
 y_s=\eta g(y'_{st},y'_{so})
 \label{Eq:HecSimple}
\end{equation}
where $\eta$  represents a constant for bridge $N$, and $g$ is a function of the stage and sonar readings.

Meanwhile, the AI forecast model using LSTM can be written as:
\begin{equation}
\centering
  \Big\{\Big[\begin{matrix}
   y^i_{st} \\
   y^i_{so} 
  \end{matrix}\Big]\Big\}^k_{i=n}
  =\psi\Big(\Big\{\Big[\begin{matrix}
   y^i_{st} \\
   y^i_{so} 
  \end{matrix}\Big]\Big\}^{n-1}_{i=m},\textbf{W},\textbf{b}\Big)   
\end{equation}
where, $\Big\{\Big[\begin{matrix}
   y^i_{st} \\
   y^i_{so} 
  \end{matrix}\Big]\Big\}^{n-1}_{i=m}$ are the sensor readings in the past $\{t^i\}^{n-1}_{i=m}=\{t^m,\cdots, t^{n-1}\}$ which are the input features for the LSTM model and \textbf{W} and \textbf{b}\ are the matrix of weights and biases of the network, which get adjusted during the training process. Therefore, $y_s$, i.e. the maximum scour depth in the forecast window of \textbf{t} can be obtained from a function like $\chi$, representing the AI surrogate function that can approximate the scour empirical equation presented in Eq.\ref{Eq:HecSimple}:
  
  \begin{eqnarray}
          y_s&=&\max(y^i_{st}-y^i_{so})^k_{i=n}-y_1\\
          &=&\chi\Big[\psi\Big(\Big[\begin{matrix}
            y^i_{st} \\
            y^i_{so} 
            \end{matrix}\Big]^{n-1}_{i=m},\textbf{W},\textbf{b}\Big)\Big] \approx \eta g(y'_{st},y'_{so})
          \label{Eq:AIFunction}
  \end{eqnarray}.
  
\section{Results}

\subsection{LSTM Performance Comparison}
\begin{figure}[H]
    \centering
    \includegraphics[width=0.5\textwidth]{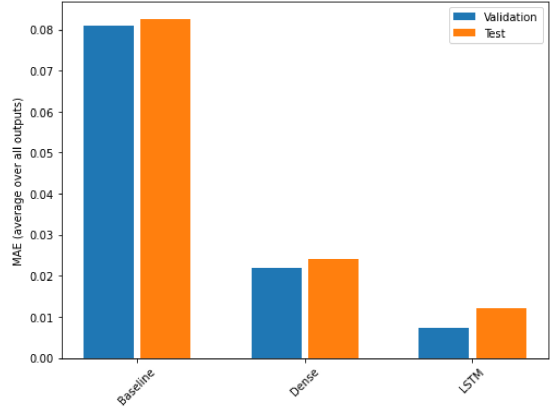}
    \caption{Comparison of Baseline, Dense, and LSTM models (MAE is normalized)}
    \label{fig: BaDeLSTM}
\end{figure}

A preliminary analysis was done to better demonstrate the LSTM capability over a baseline linear model and a Dense model. The Dense model includes only a single Dense layer with ReLU activation function. The baseline model was defined as a model that at each time-step predicts the same value as the last time-step. %Input-label width of 24-24 was considered for data slicing and 
The performance of the models was compared based on MAE over the test dataset on a smaller fraction of the data for bridge 742. Figure \ref{fig: BaDeLSTM} shows that LSTM significantly outperforms the baseline and the Dense model, demonstrating LSTM superiority in forecast compared to the other models. 

\subsection{Model Optimization and Hyper-Parameter Tuning}
\begin{figure}[H]
   \centering
     \begin{subfigure}{\textwidth}
     \centering
      \includegraphics[width=0.9\textwidth]{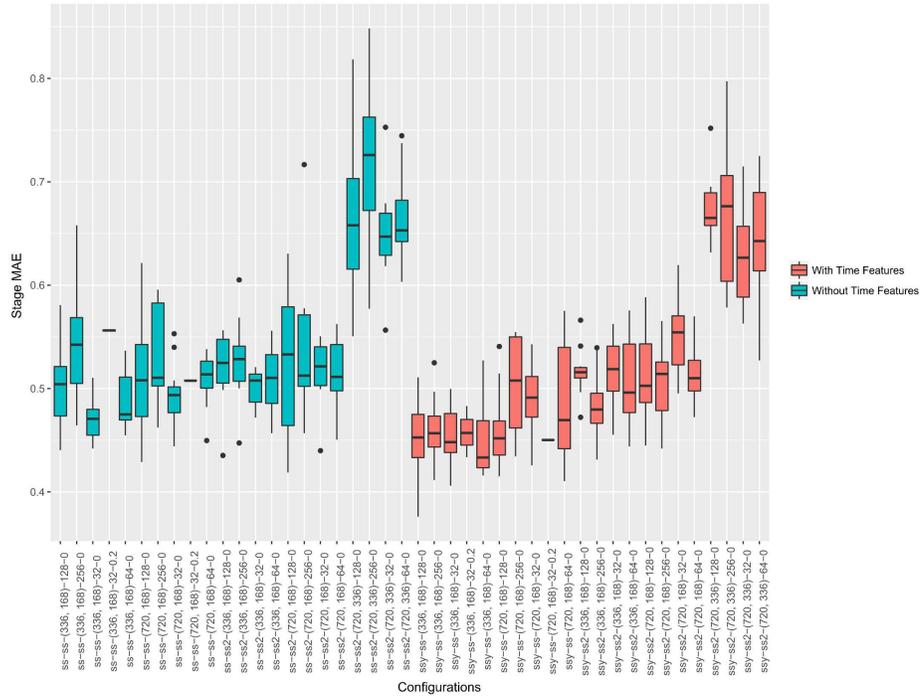}
      \caption{}
     \end{subfigure}
     \begin{subfigure}{\textwidth}
     \centering
      \includegraphics[width=0.9\textwidth]{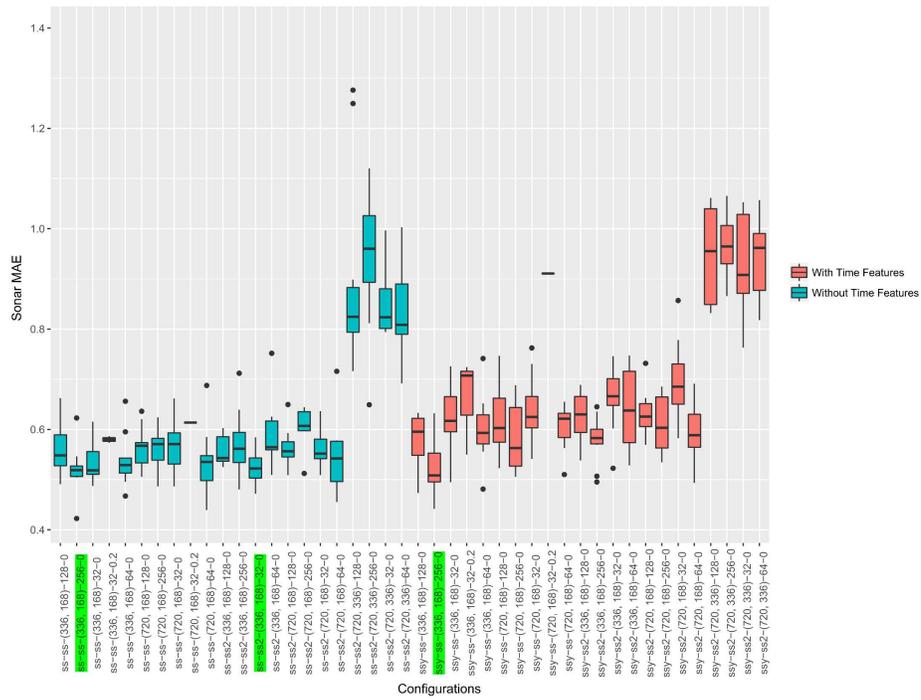}
      \caption{}
     \end{subfigure}
  \caption{Grid-search results showing, a) stage MAE and b) sonar MAE boxplots for various configurations, with the top three models highlighted - bridge 742 (MAE is normalized)}
  \label{fig: 742-mae}
\end{figure}

\begin{figure}[H]
   \centering
     \begin{subfigure}{\textwidth}
     \centering
      \includegraphics[width=0.9\textwidth]{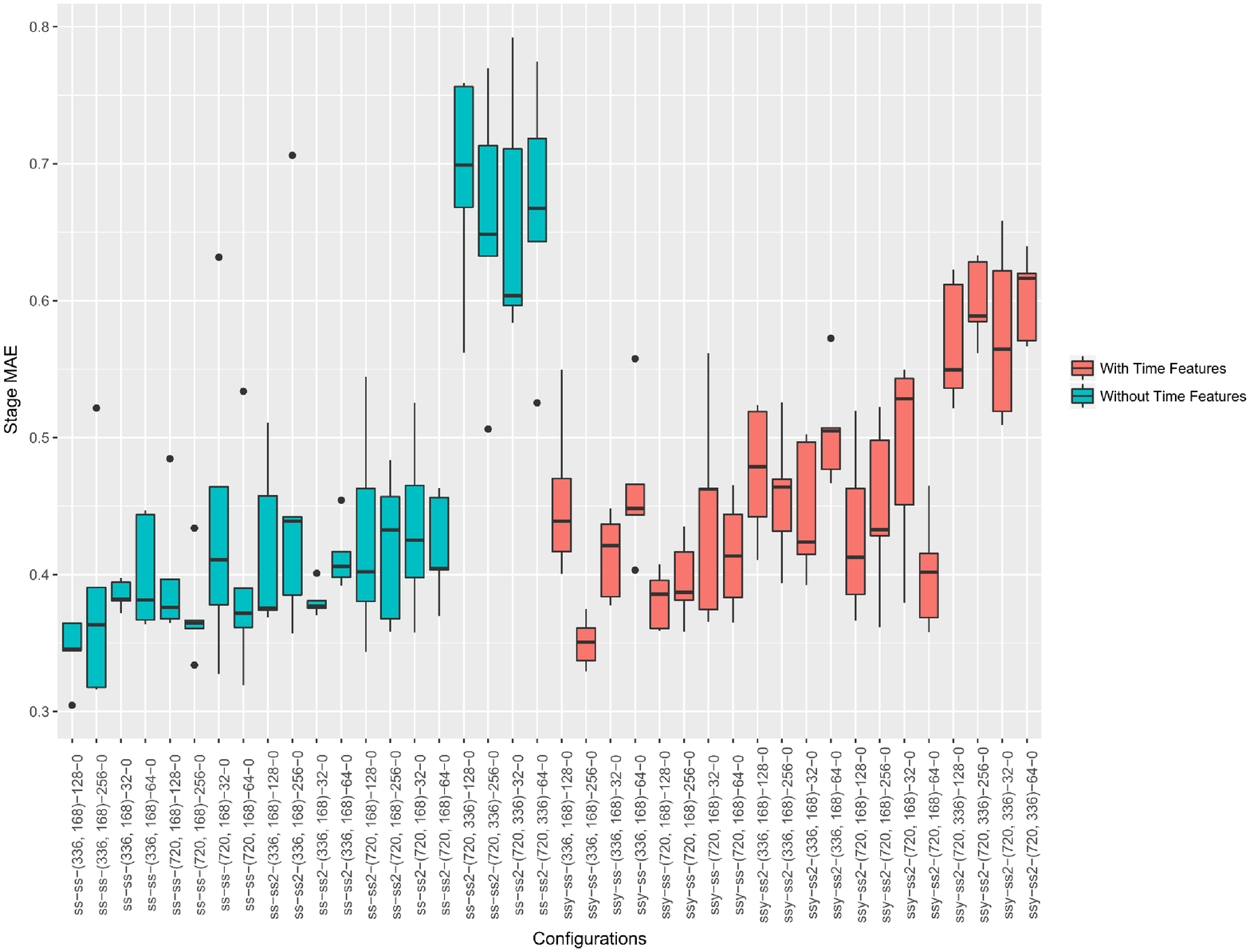}
      \caption{}
     \end{subfigure}
     \begin{subfigure}{\textwidth}
     \centering
      \includegraphics[width=0.9\textwidth]{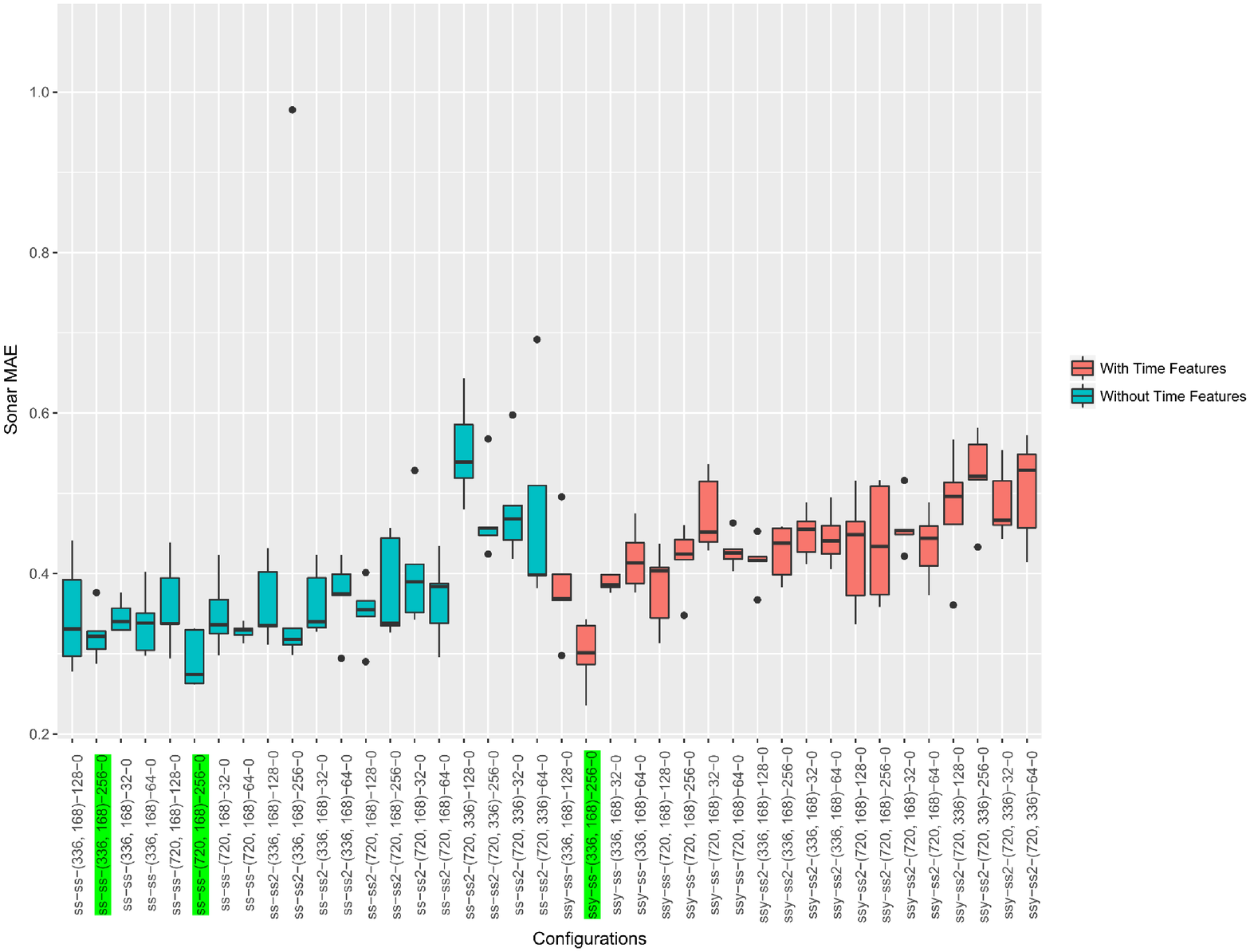}
      \caption{}
     \end{subfigure}
  \caption{Grid-search results showing, a) stage MAE and b) sonar MAE boxplots for various configurations, with the top three models highlighted - bridge 539 (MAE is normalized)}
  \label{fig: 539-mae}
\end{figure}

\begin{figure}[H]
   \centering
     \begin{subfigure}{\textwidth}
     \centering
      \includegraphics[width=0.9\textwidth]{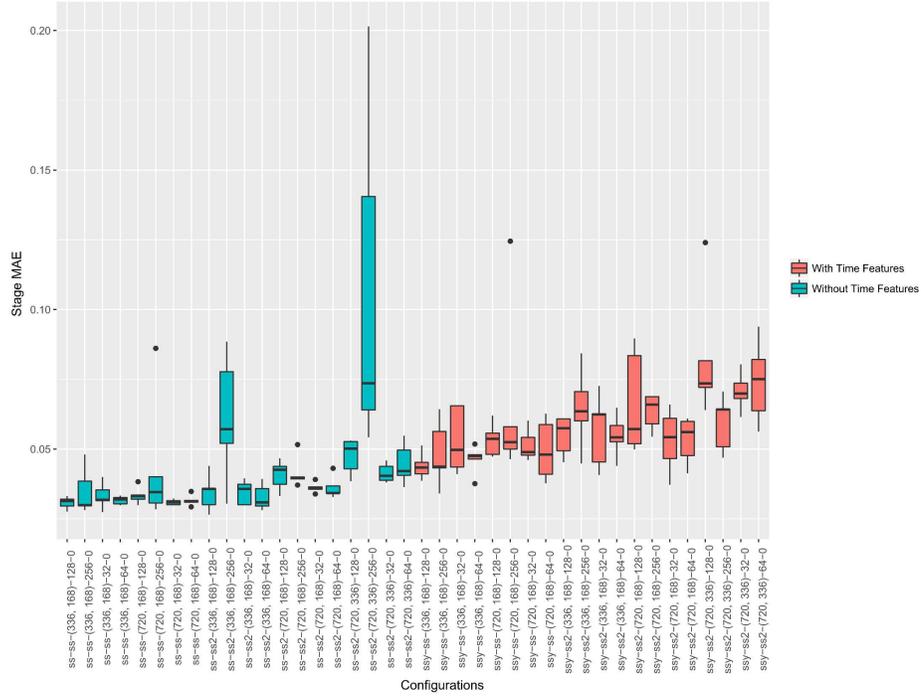}
      \caption{}
     \end{subfigure}
     \begin{subfigure}{\textwidth}
     \centering
      \includegraphics[width=0.9\textwidth]{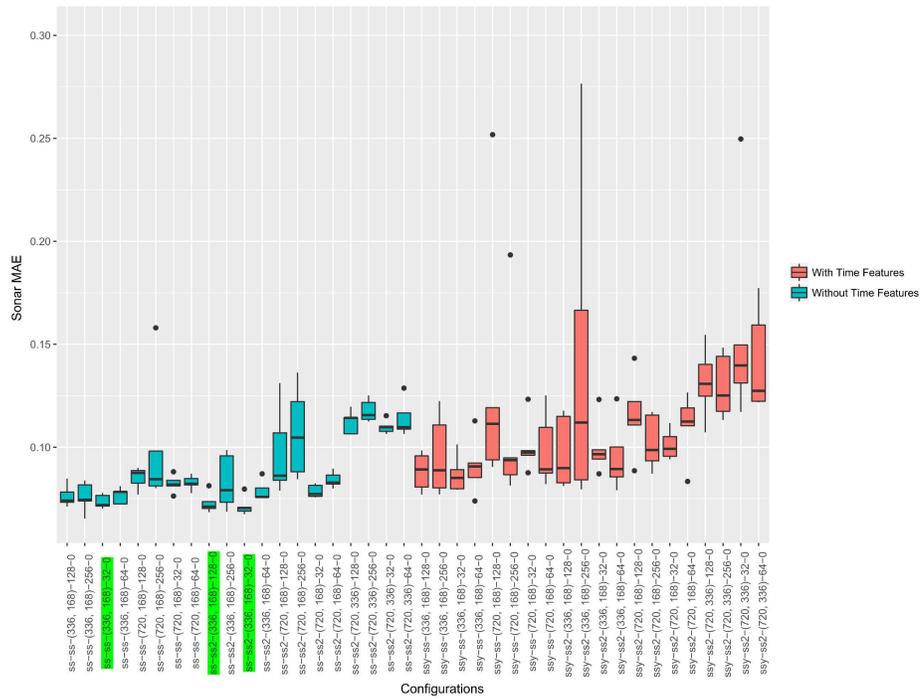}
      \caption{}
     \end{subfigure}
  \caption{Grid-search results showing, a) stage MAE and b) sonar MAE boxplots for various configurations, with the top three models highlighted - bridge 230 (MAE is normalized)}
  \label{fig: 230-mae}
\end{figure}

Grid-search analyses were computationally-intensive; however, the computational time was significantly reduced by running the codes on the GPU clusters. Each training epoch took about 400-500 seconds on two GPU node. For each configuration, the LSTM network was retrained from scratch for a large number of times to evaluate the uncertainty in the training process. Figure \ref{fig: 742-mae} to \ref{fig: 230-mae} provide the box-plots of MAE values for various LSTM configurations. Note that the MAE values are based on normalized data (therefore unitless). Configurations were encoded as follows: 

\begin{center}
\texttt{[features]-[model]-[window]-[units]-[drop-out]}
\end{center}

For example, \texttt{ssy-ss-(336,168)-256-0} corresponds to the configuration that includes time features, uses the Single-Shot model, has 336 time-steps as input and 168 time-steps for labels, includes 256 units, and does not have drop-out regularisation. 
%The abbreviation \texttt{ssy} under \textttt{[features]} corresponds to the combination with time features and \texttt{ss} corresponds to the combination with no time features. Under \textttt{[model]}, \texttt{ss} corresponds to the Single Shot model, \texttt{ss2} corresponds to the Two-Layer, Single-Shot model, and \texttt{fd} represents the Feed-Back model. %
The best configurations were selected based on the smallest MAE mean and standard deviation (see Table \ref{tab: hyperparams_values} for selected hyper-parameter values). 

The following observations were made from these analyses:

\begin{itemize}
  \item Adding time features did not improve the LSTM predictions of bed elevation, yet their presence showed to slightly improve the predictions of river stage for bridge 742.
  \item Increasing the input width from 336 hrs (14 days) to 720 hrs (30 days) while keeping the label width (forecast window) constant did not improve the prediction accuracy and showed to amplify the variability in model predictions.
  \item Increasing the label width (forecast window) from 7 days to 14 days showed to consistently increase the average prediction error.
  \item Increasing the LSTM layers to two didn't show a meaningful improvement in the accuracy of predictions but increased the variability of predictions. This may be due to over-fitting, where the more complex model (more nonlinear units) learns better on the training data but fails to generalize on the test data.
  \item The Feedback model did not outperform the Single-Shot model. This could be related to the error cascade effect from the initial steps within the label sequence to the next time steps. As the analyses have been performed on a limited number of bridges, the Feedback model may outperform the Single-Shot model in other bridges, depending on the training data.
  \item Using the drop-out method (0.2 ratio) did not improve the accuracy of predictions.
  \item The optimum number of units showed to vary from bridge to bridge and dependant on the other hyper-parameters of the model (e.g., feature combination, model variant, and architecture).
\end{itemize}

\subsection{Training}
\begin{figure}[H]
 \centering
    \includegraphics[width=0.9\textwidth]{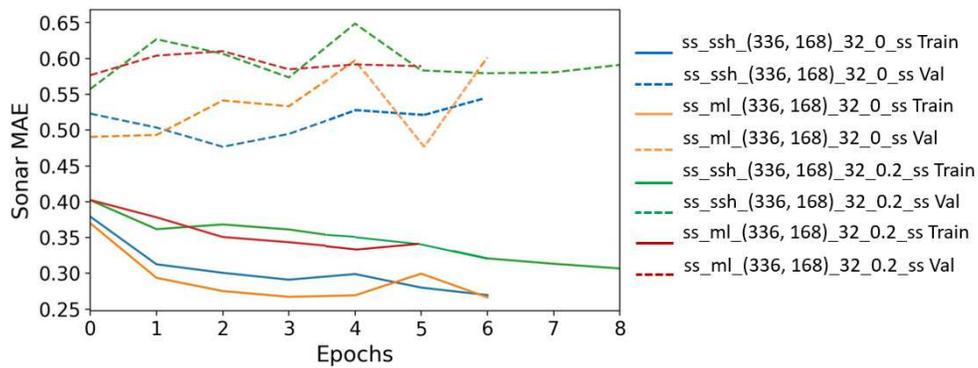}
    \caption{Training history plots: variation of the MAE for sonar prediction with epochs for training and validation datasets}
     \label{fig: error-epochs}
\end{figure}

\begin{figure}[H]
 \centering
    \includegraphics[width=0.45\textwidth]{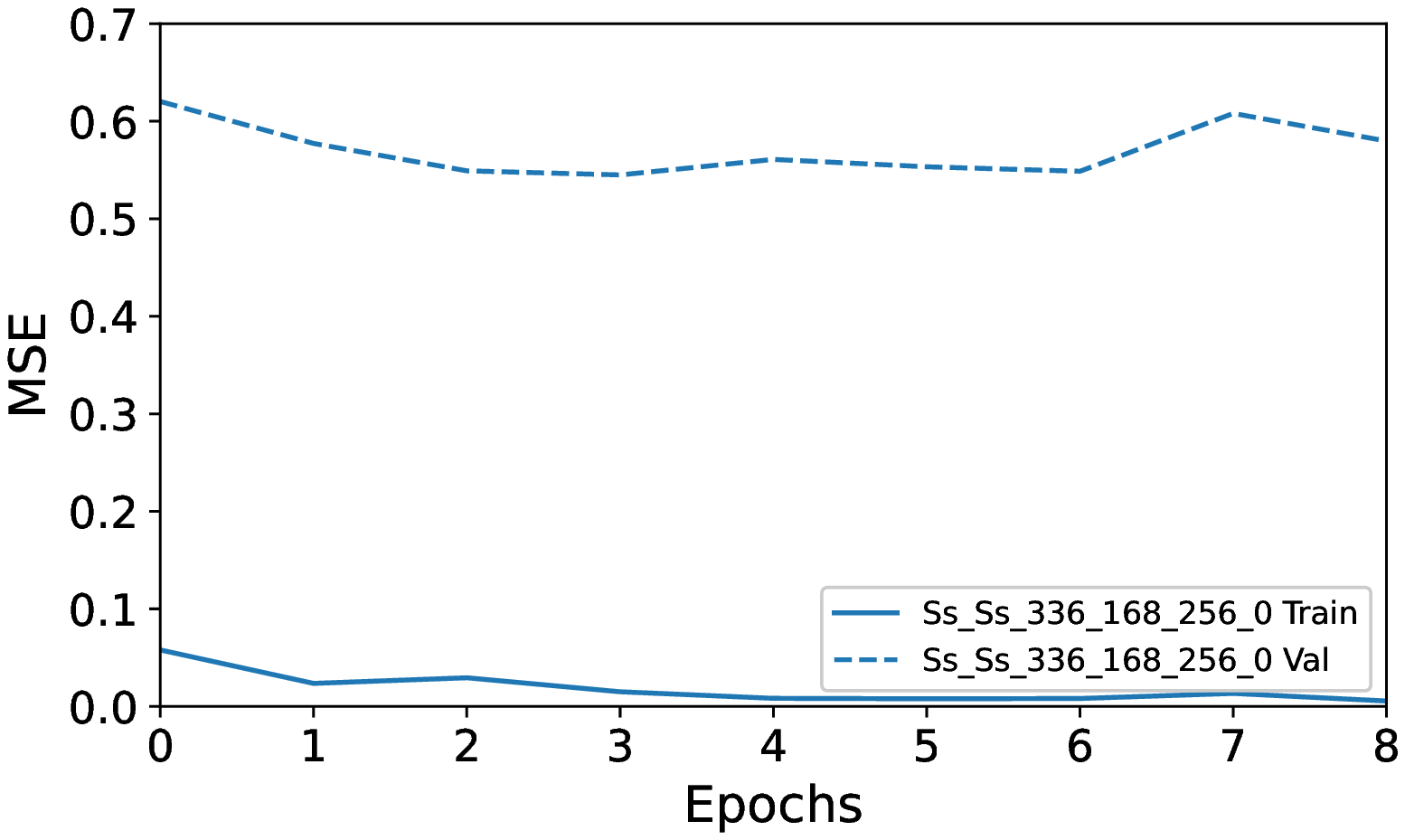}
    \includegraphics[width=0.45\textwidth]{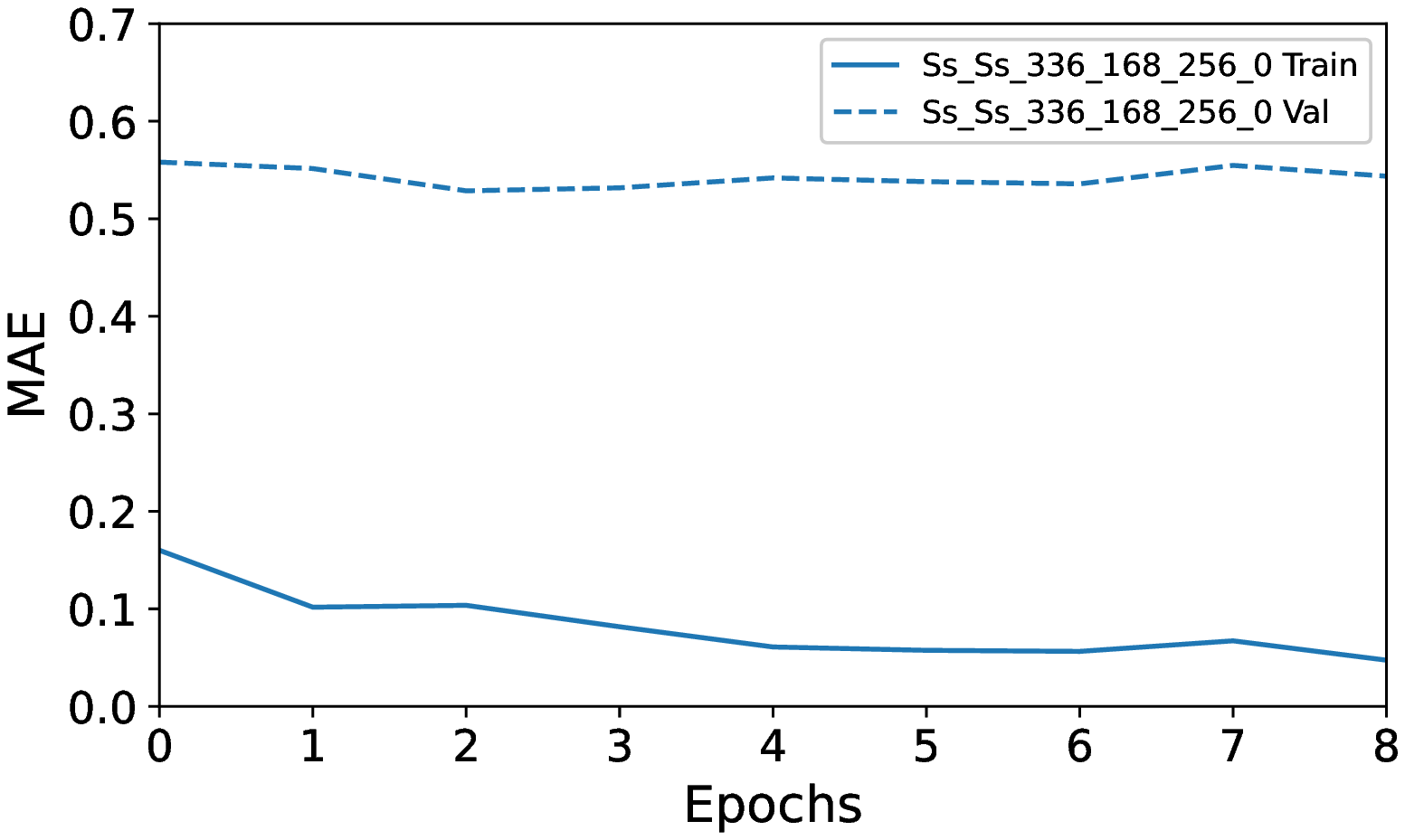}
    \caption{Comparison between MSE and MAE variations with number of epochs during training} 
    \label{fig: MSEvsMAE}
\end{figure}

Figure \ref{fig: error-epochs} provides the graphs of the training process for a number of LSTM configurations, showing the variation of MAE in training and validation versus the number of epochs. As mentioned earlier, the stopping criteria was applied based on the error progression over validation dataset to avoid over-fitting. Note that the MAE is based on the standard-normalized values. 
Figure \ref{fig: MSEvsMAE} provides a comparison between the two evaluation metrics, loss(MSE) and MAE across training epochs. Both metrics show consistency in their trends across epochs. The main difference between these two metrics is that, loss (\textit{MSE}) is calculated based on the error on the vector of features, i.e. both sonar and stage predictions, whereas \textit{MAE} is calculated separately for sonar and stage to better evaluate of scour prediction accuracy. 

\subsection{Prediction Results}
The AI forecast of river stage and bed elevation variation for the three case study bridges are presented and discussed in this section. The forecast results are shown for the test dataset, which is the last year of bridge monitoring data. The results pertain to the most optimum model selected through the grid-search analyses, as explained in the previous section. 

\subsubsection{Chilkat River Bridge - 742} 
\begin{figure}[H]
  \begin{subfigure}{\textwidth}
    \centering
    \includegraphics[width=0.8\textwidth]{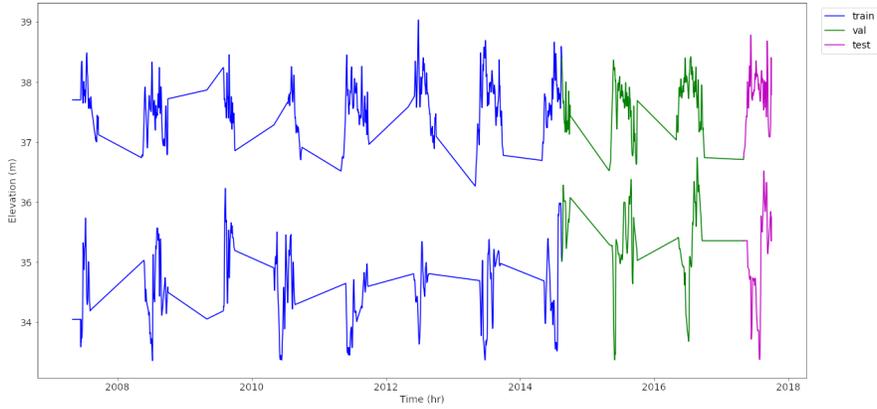}
    \caption{}
    \label{fig: datadiv-742}
  \end{subfigure}
  \begin{subfigure}{\textwidth}
   \centering
   \includegraphics[width=0.7\textwidth]{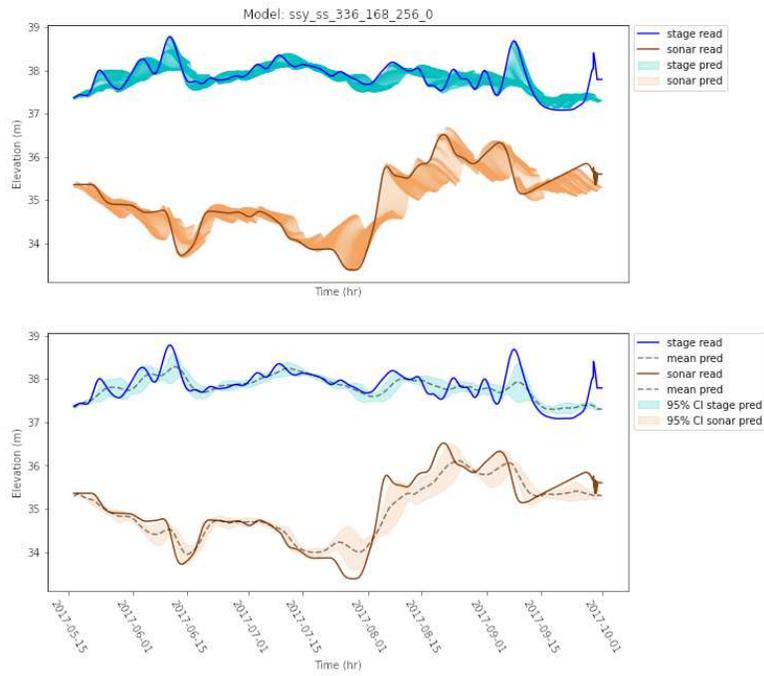}
   \caption{}
   \label{fig: pred-742}
  \end{subfigure} 
  \caption{a) Data partitioning for AI training, b) AI predictions, mean, and confidence intervals over the test dataset- bridge 742.}
\end{figure}

Figure \ref{fig: datadiv-742} shows the data division for training, validation, and test datasets. Figure \ref{fig: pred-742} presents bed elevation and river stage predictions throughout 2017 (test dataset). The top graph shows the processed sensor readings and a shade created by the prediction of the time-steps for every data sequence over the label width. For bridge 742, the shade shows the predictions of 168 time-steps (7 days) into the future for every 336 time-steps of history (14 days), each of which is shifted to the right by one time-step. The shades closely follow the trends of the actual readings for both sonar and stage, showing the model's effectiveness for scour forecast. On the bottom part, besides the actual readings, the average and 95\% confidence interval of the predictions are shown. The mean absolute error of bed elevation (sonar) predictions over the test dataset is 0.19m. Also the variability of predictions is within acceptable limits, with a standard deviation less than 0.3m (1ft).

As mentioned earlier, bridge 742 shows a seasonal variation of bed elevation of up to 3m, including about 2m scour from June to August and subsequent 3m of filling from August to October. The LSTM model shows to successfully follow the seasonal trend of scour and filling throughout 2017, having learned from long-term patterns in the training data from 2007 through 2016. During the scour episodes, the lower bound predictions consistently provide an accurate assessment of bed elevation variation, and upper bound predictions closely track the bed elevation variations during filling episodes. Using the lower bound for scour and upper bound for filling episodes, the AI predictions show to provide an accurate conservative estimate of the bed elevation (hence scour assessment) at any given time.

\subsubsection{Knik River Bridge - 539}
\begin{figure}[H]
  \begin{subfigure}{\textwidth}
    \centering
    \includegraphics[width=0.8\textwidth]{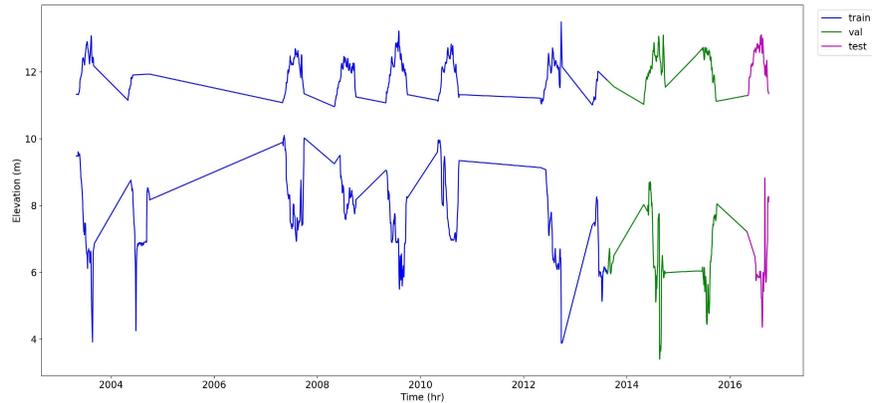}
    \caption{}
    \label{fig: datadiv-539}
  \end{subfigure}
  \begin{subfigure}{\textwidth}
   \centering
   \includegraphics[width=0.7\textwidth]{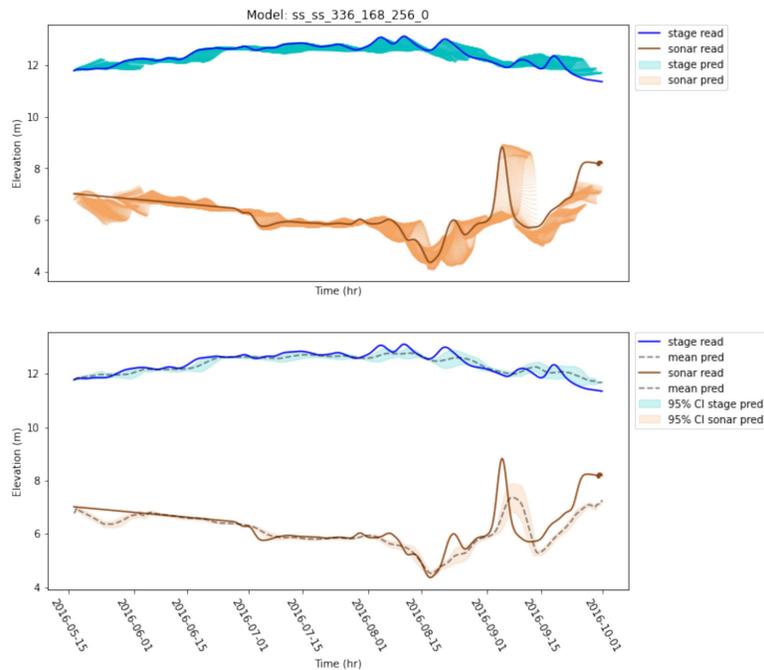}
   \caption{}
   \label{fig: pred-539}
  \end{subfigure} 
  \caption{a) Data partitioning for AI training, b) AI predictions, mean, and confidence intervals over the test dataset - bridge 539.}
\end{figure}

The data division for training is shown in Figure \ref{fig: datadiv-539}. The bed elevation and river stage predictions throughout 2016 (test dataset) is shown in Figure \ref{fig: pred-539}. Similar to bridge 742, the forecast is performed 7 days (168hrs) in advance with 14 days (336hrs) of history. The shades of LSTM predictions closely follow the trend of the actual sonar and stage readings. The mean absolute error of bed elevation prediction is about 0.25m over the test dataset. The variability of predictions is within acceptable limits with a standard deviation less than 0.3m (1ft).

Similar to bridge 742, the AI predictions successfully follow the seasonal scour and filling trends throughout 2016, even during the acute changes between August to September with about 4m variation of bed elevation. The variability in bed elevation predictions is small, with a maximum standard deviation below half a meter during the pick of filling in September 2016.

\subsubsection {Sheridan Glacier Bridge - 230}
\begin{figure}[H]
  \begin{subfigure}{\textwidth}
    \centering
    \includegraphics[width=0.8\textwidth]{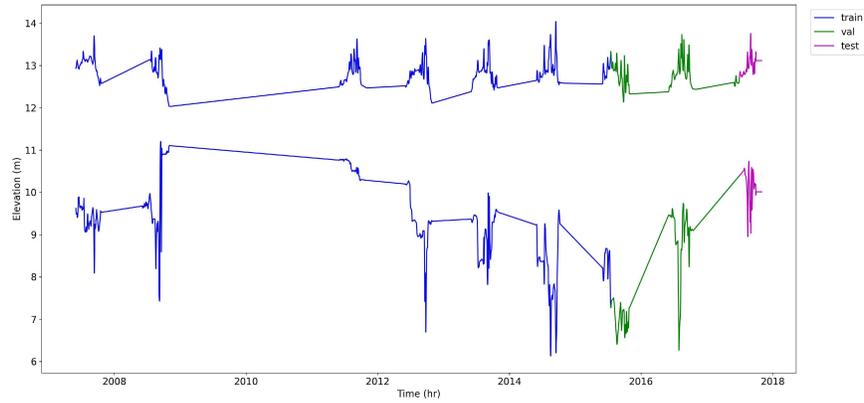}
    \caption{}
    \label{fig: datadiv-230}
  \end{subfigure}
  \begin{subfigure}{\textwidth}
   \centering
   \includegraphics[width=0.7\textwidth]{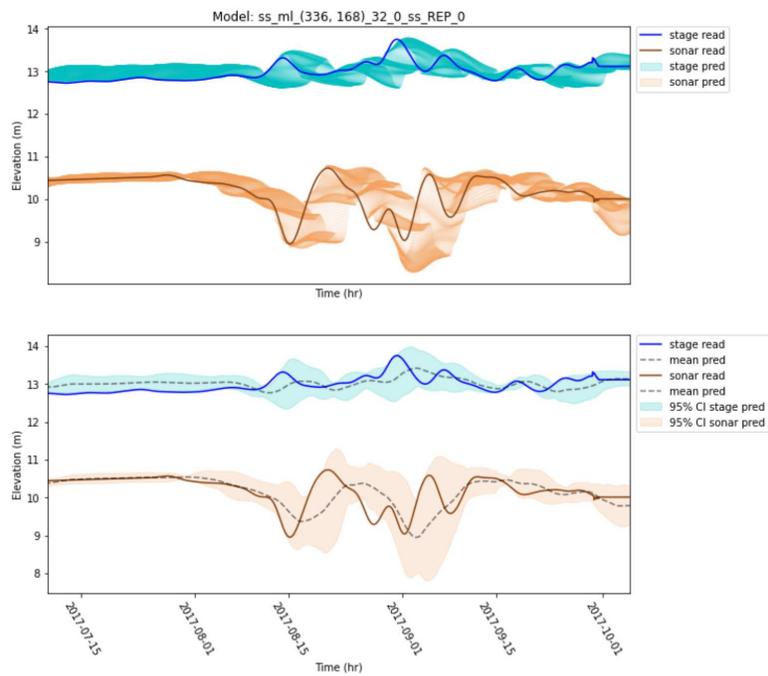}
   \caption{}
   \label{fig: pred-230}
  \end{subfigure} 
  \caption{a) Data partitioning for AI training, b) AI predictions, mean, and confidence intervals over the test dataset - bridge 230.}
\end{figure}

The data division and AI predictions for bridge 230 are shown in Figure \ref{fig: datadiv-230} and \ref{fig: pred-230}. More variability in bed elevation and river stage predictions is observed for bridge 230 compared to the other two bridges; nevertheless the trend (mean) of predictions follows the scour and filling episodes throughout 2017 (test dataset). This includes consecutive scour and filling episodes happening between August and September, with a maximum bed elevation variation of about 2m (6.5ft). Similar to the other two bridges, the forecast is performed 7 days (168hrs) in advance with 14 days (336hrs) of history. The mean absolute error of bed elevation prediction is less than 0.37m over the test dataset. It is worth noting that although the mean of prediction provides a reasonable forecast of scour at the bridge pier, the higher variability in predictions have resulted in the lower bound to indicate larger (more conservative) scour depth in some episodes, for e.g., around end of August and mid-September 2017. 

\subsection{Discussion on the Performance of the LSTM Models}
Table \ref{Tab: LSTM summary} summarizes the performance of the selected LSTM models for each bridge based on MAE for sonar and stage predictions. The average prediction error for bed elevation varies between 0.2 to 0.4m and for stage water between 0.1 to 0.2m. The average scour error percentage (= sonar MAE / max depth of scour) is less than 10\% for bridge 742 and 539 but jumps higher to 25\% for bridge 230. The reason for this larger error lies with a considerable amount of missing data (year 2008 and 2009) for bridge 230, as opposed to the other bridges.
To provide a comparison basis with best of the empirical models, according to Sheppard et a.~\citeyear{Sheppard:2014}, HEC-18 mean absolute error percentage over a substantially large lab dataset has been estimated around 21\%.  

\begin{figure}
\centering
 \begin{subfigure}{\textwidth}
 \centering
  \includegraphics[width=0.8\textwidth]{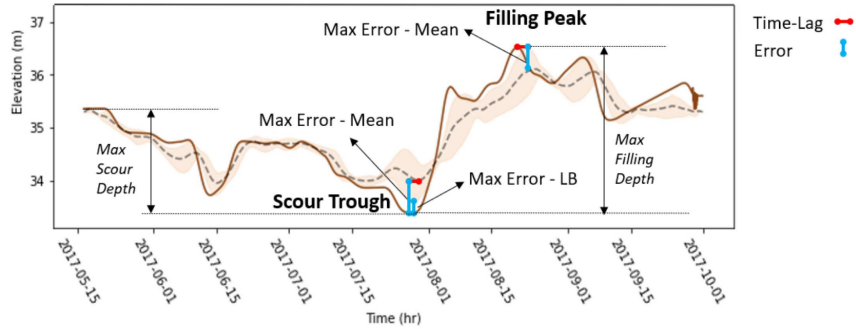}
   \caption{} 
 \end{subfigure}
 \begin{subfigure}{\textwidth}
 \centering
  \includegraphics[width=0.8\textwidth]{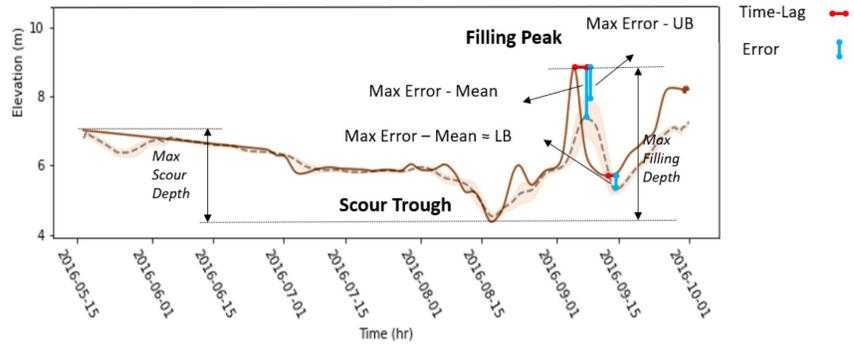}
   \caption{} 
 \end{subfigure}  
 \begin{subfigure}{\textwidth}
 \centering
   \includegraphics[width=0.6\textwidth]{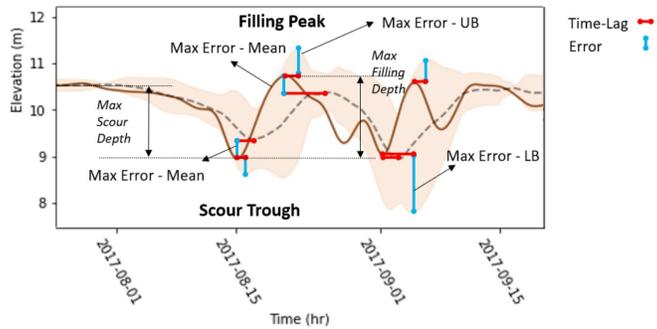}
   \caption{}
 \end{subfigure}  \caption{Definition of maximum errors based on mean and LB/UB predictions for scour troughs and filling peaks: a) bridge 742, b) bridge 539, c) bridge 230}
\label{fig: ScourFilling}
\end{figure}

The maximum error in scour trough and filling peak forecasts are provided in Table \ref{tab: ScourTroughPerf} and graphically shown in Figure \ref{fig: ScourFilling}. The maximum error based on the mean of predictions varies between 0.5m to 0.7m for scour troughs and 0.4m to 1.7m for filling peaks. The lower bound (LB) and upper bound (UB) errors show a reasonable degree of variability in the LSTM predictions, varying between 0.2m to 0.9m for scour, and 0m to 1.4m for filling.

\begin{table}
\caption{Performance summary of the LSTM models over test dataset}
    \centering
    \begin{tabular}{c|c|c|c|c}
      \textbf{Bridge} &\textbf{Sonar MAE-m} &\textbf{Stage MAE-m} &\textbf{Max Scour Depth-m} &\textbf{Scour Error-\%}\\
      \hline
      \hline
      742 & 0.19 & 0.19 & 2.1 & 9\\	
      539 & 0.25 & 0.13 & 3.3 & 7.5\\\	
      230 & 0.37 & 0.08 & 1.5 & 25\\
    \end{tabular}
\label{Tab: LSTM summary}
\end{table}

\begin{table}
\caption{Maximum prediction error for scour (troughs) and filling (peaks) across the test dataset}
    \begin{tabular}{c|c|c|c|c}
     \centering
     \textbf{Bridge} &\textbf{Mean Error (m)} &\textbf{LB error (m)} &\textbf{Mean Error (m)} &\textbf{UB Error (m)}\\
     &\textbf{Scour} &\textbf{Scour} &\textbf{Filling} &\textbf{Filling}\\
      \hline
      \hline
      742 & 0.7 & 0.2 & 0.4 & 0\\
      539 & 0.6 & 0.6 & 1.7 & 0.8\\
      230 & 0.5 & 0.9 & 0.4 & 1.4\\
    \end{tabular}
\label{tab: ScourTroughPerf}
\end{table}

\subsection{Impact of Flow Velocity (Discharge)}
\label{sec: velocity}
In order to explore whether velocity is a critical feature in presence of stage time-series, we incorporated the discharge measurements (\texttt{discharge}), obtained from the USGS website, into the LSTM models for bridge 742 as an input feature and compared the performance among three different feature combinations: \texttt{ssd:[sonar, stage, discharge]}, \texttt{sd:[sonar, discharge]}, and \texttt{ss:[sonar, stage]}. Discharge is computed based on gage-height records (flow velocity) multiplied the river cross-section area. Gage-height records are obtained by systematic observation of a non-recording gage, or with automatic water level sensors relayed by remote gagging stations~\cite{USGS:2010}.

\begin{figure}[H]
   \centering
   \includegraphics[width=0.4\textwidth]{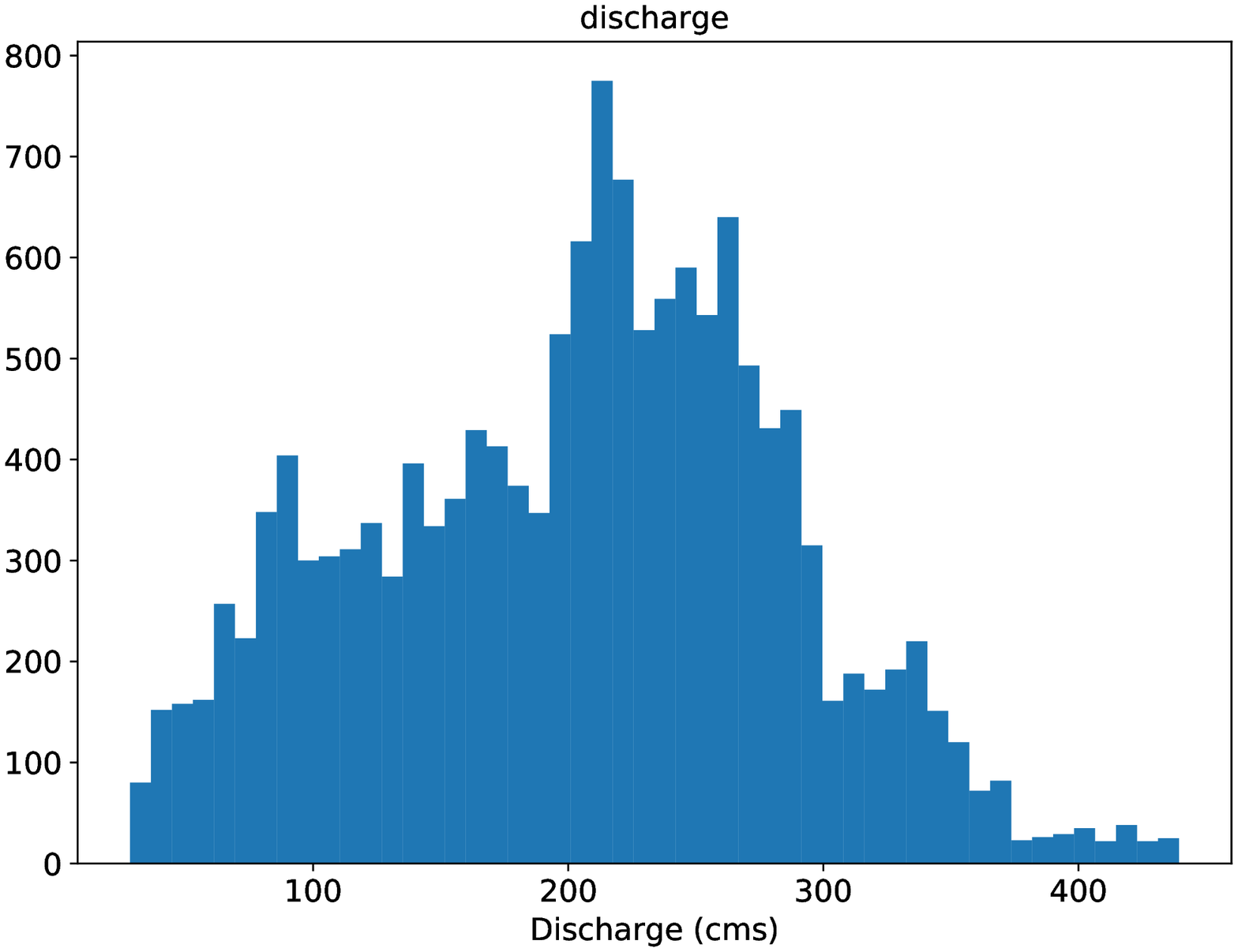}
   \includegraphics[width=0.4\textwidth]{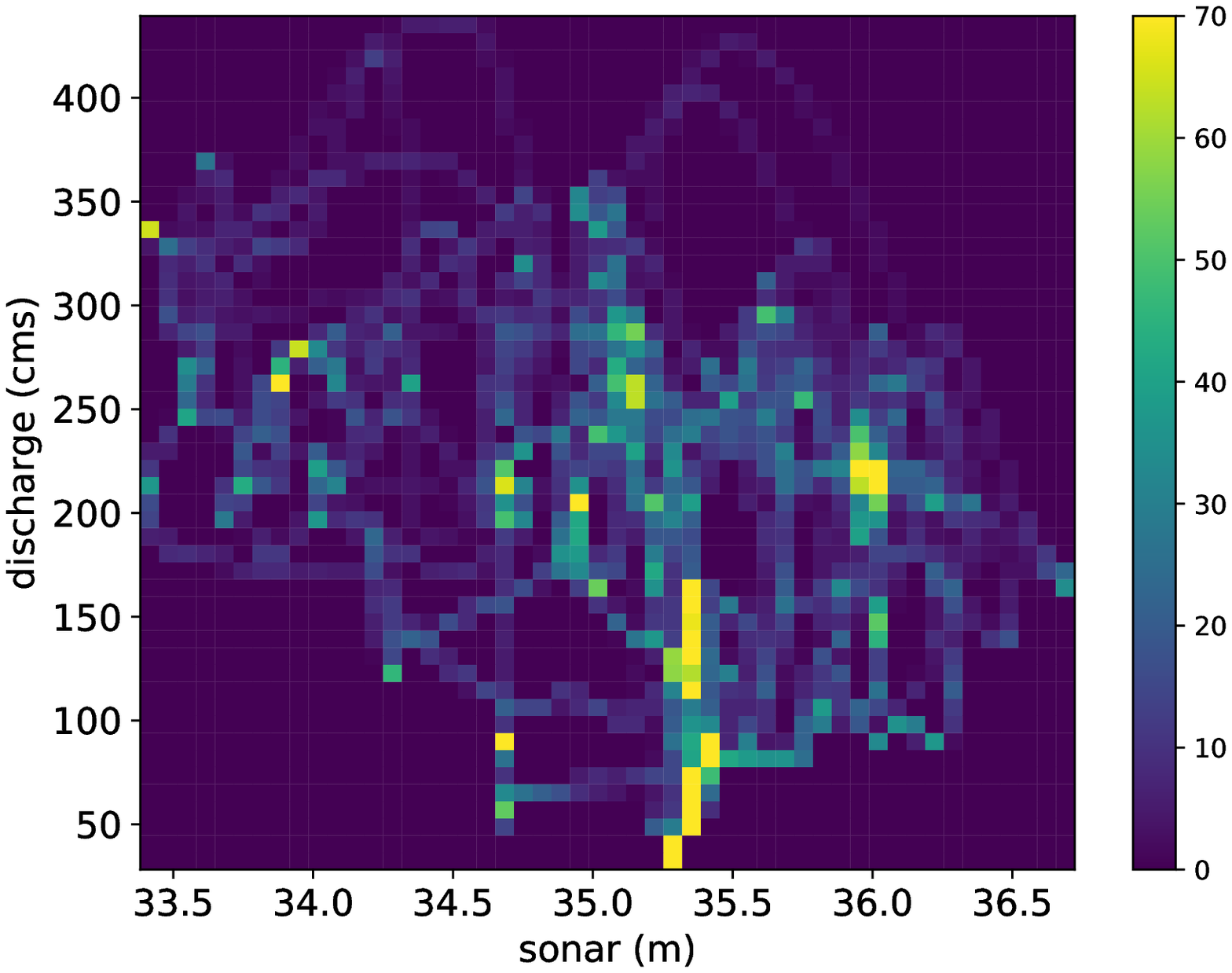}
   \includegraphics[width=0.4\textwidth]{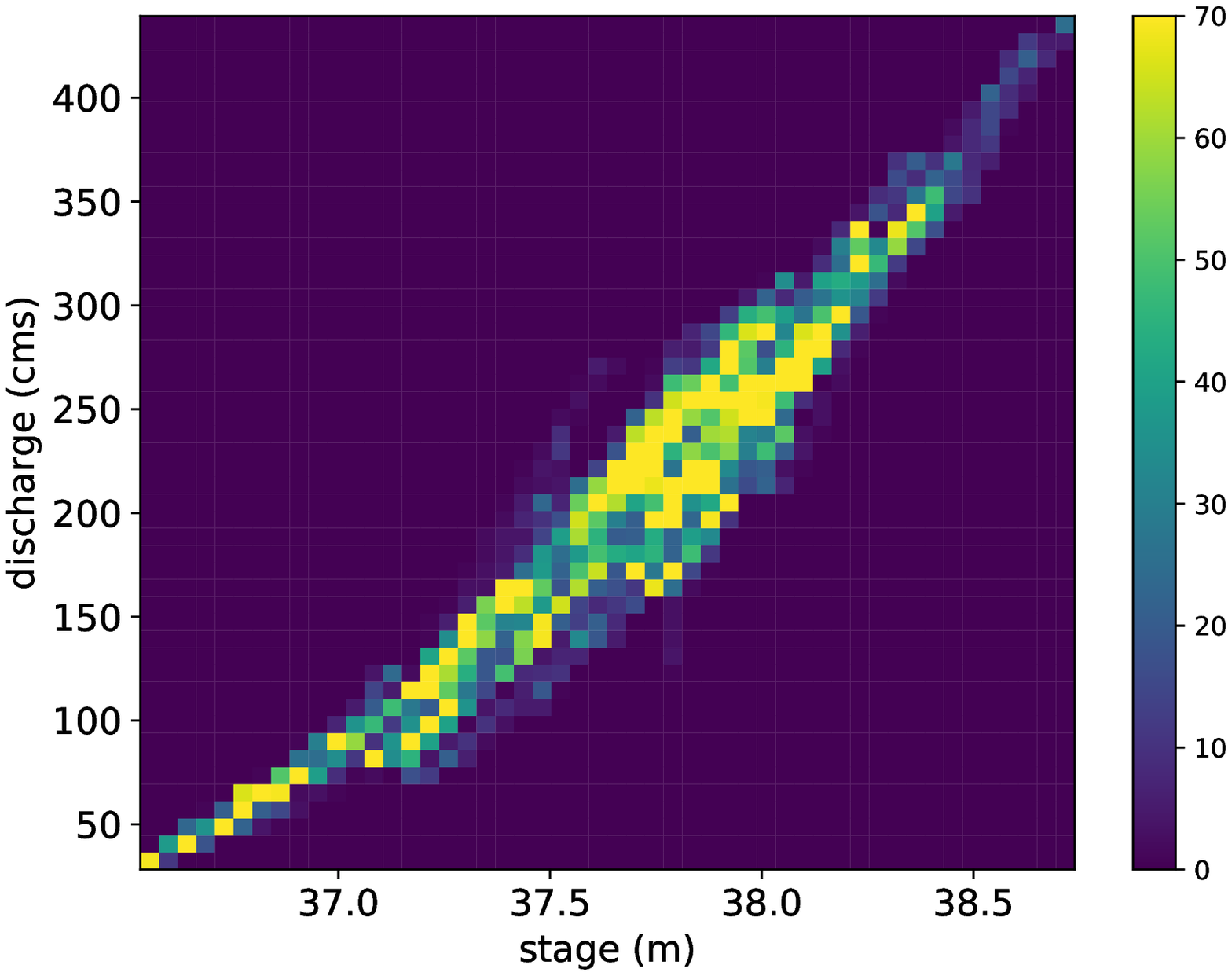}
   \caption{Histograms of discharge data - bridge 742}
   \label{fig: 742-discharge hist}
\end{figure}

\begin{figure}[H]
   \centering
   \includegraphics[width=0.5\textwidth]{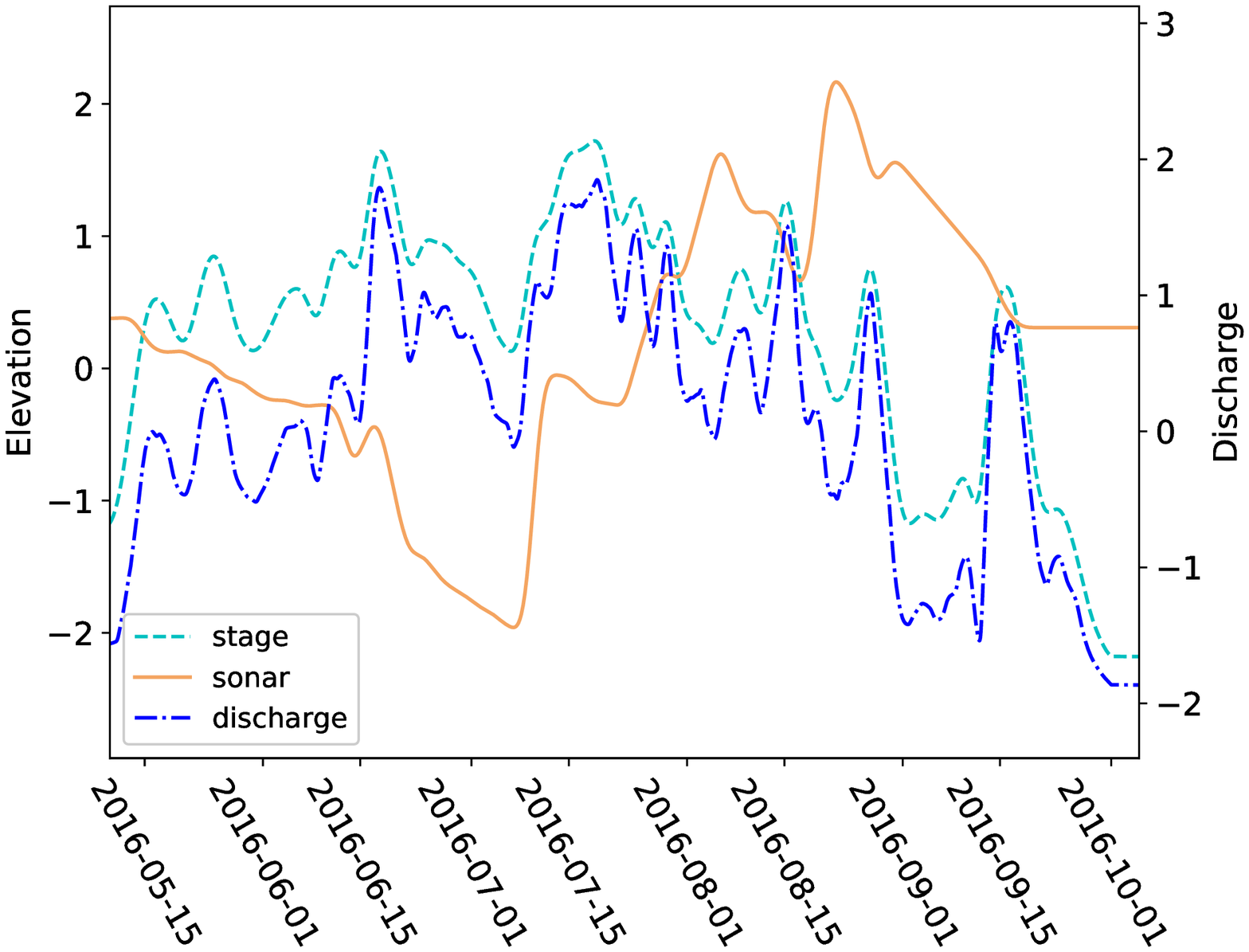}
   \includegraphics[width=0.5\textwidth]{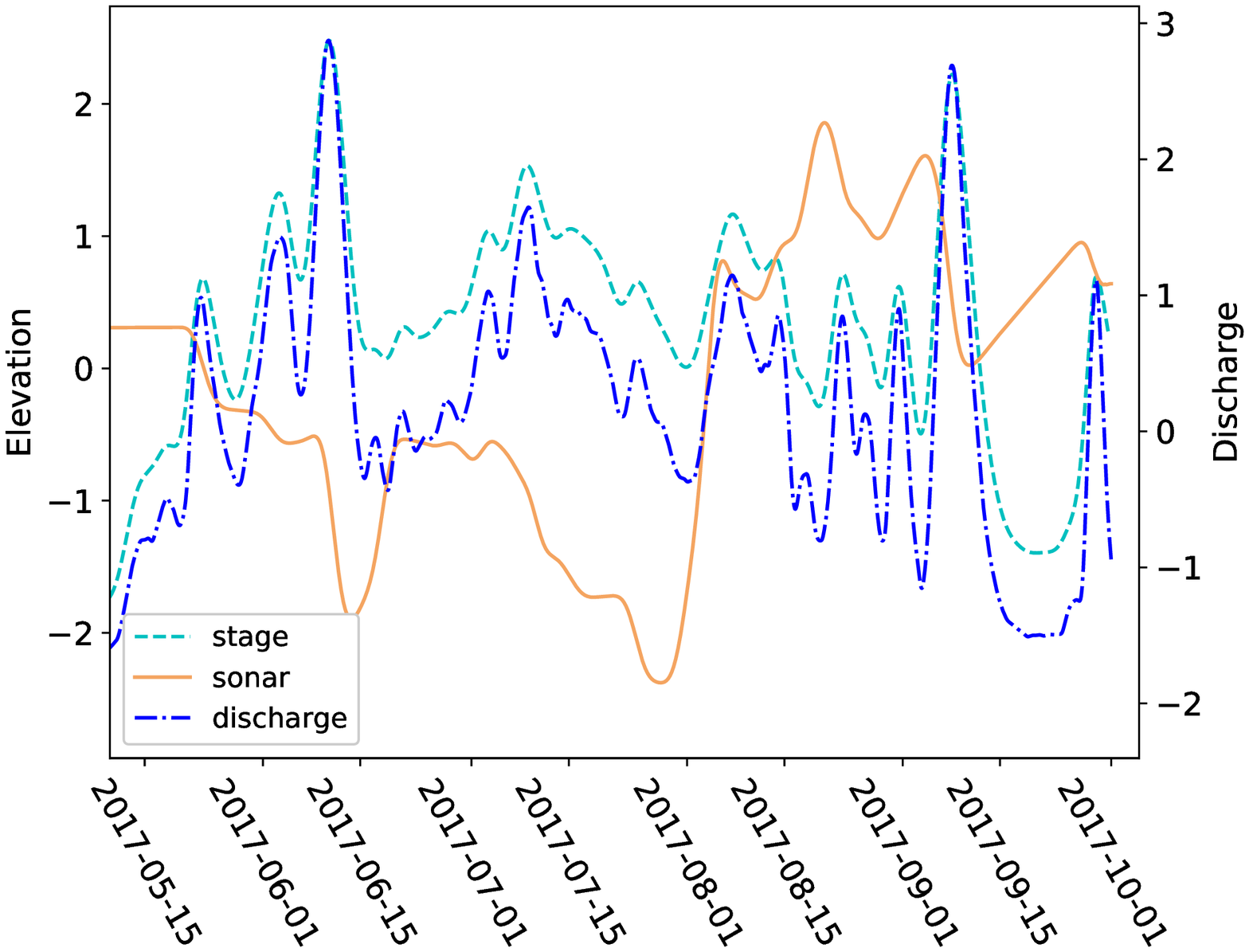}
   \caption{Sample of discharge, stage and sonar time-series - bridge 742}
  \label{fig: 742-discharge data sample}
\end{figure}

Figure \ref{fig: 742-discharge hist} provides histograms of the discharge time-series for bridge 742 and its cross-correlation with sonar and stage. Stage and discharge show a large positive correlation as observed both in Figure \ref{fig: 742-discharge hist} and Figure \ref{fig: 742-discharge data sample}.

\begin{figure}[H]
   \centering
   \includegraphics[width=0.3\textwidth]{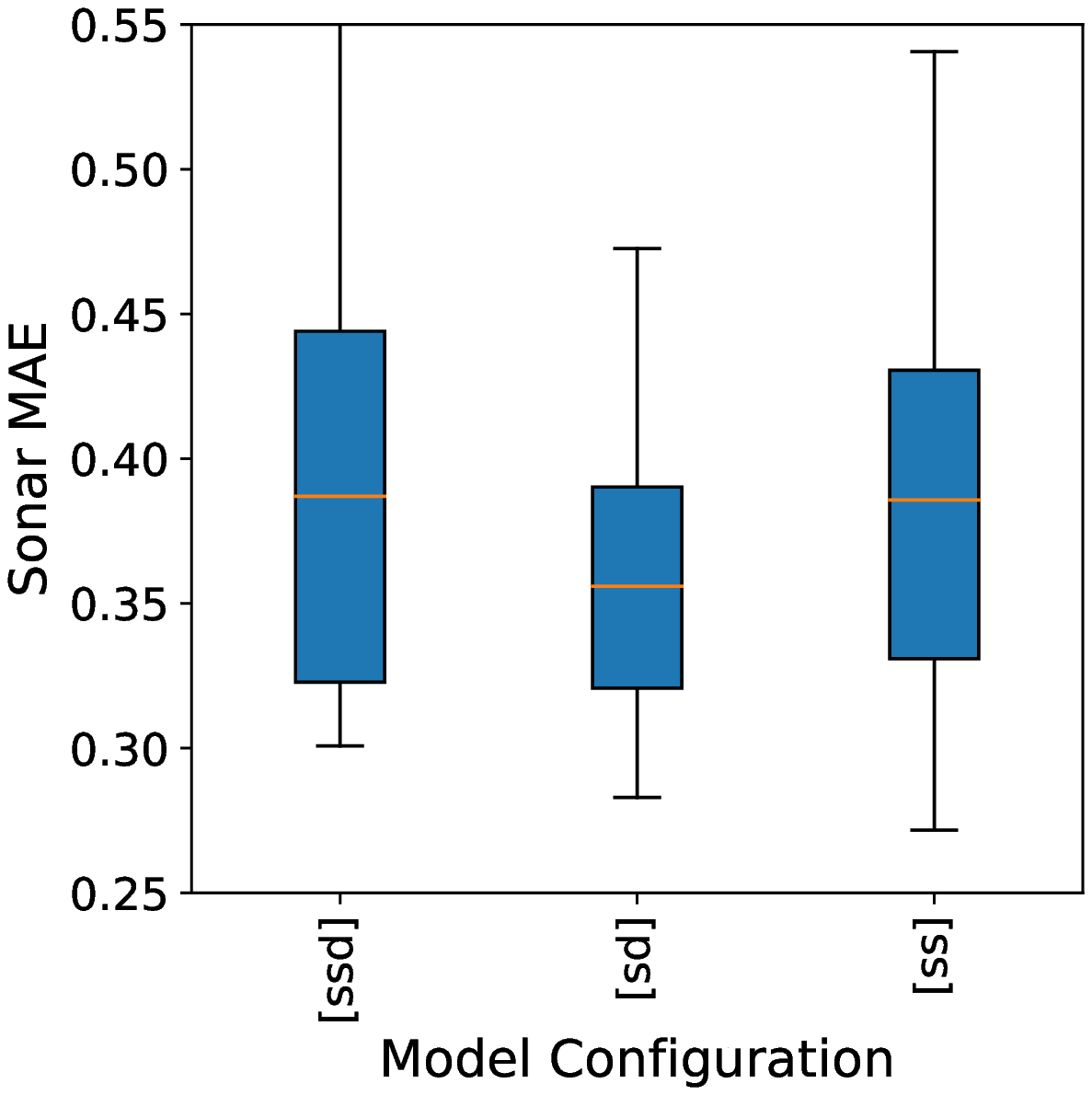}
   \includegraphics[width=0.3\textwidth]{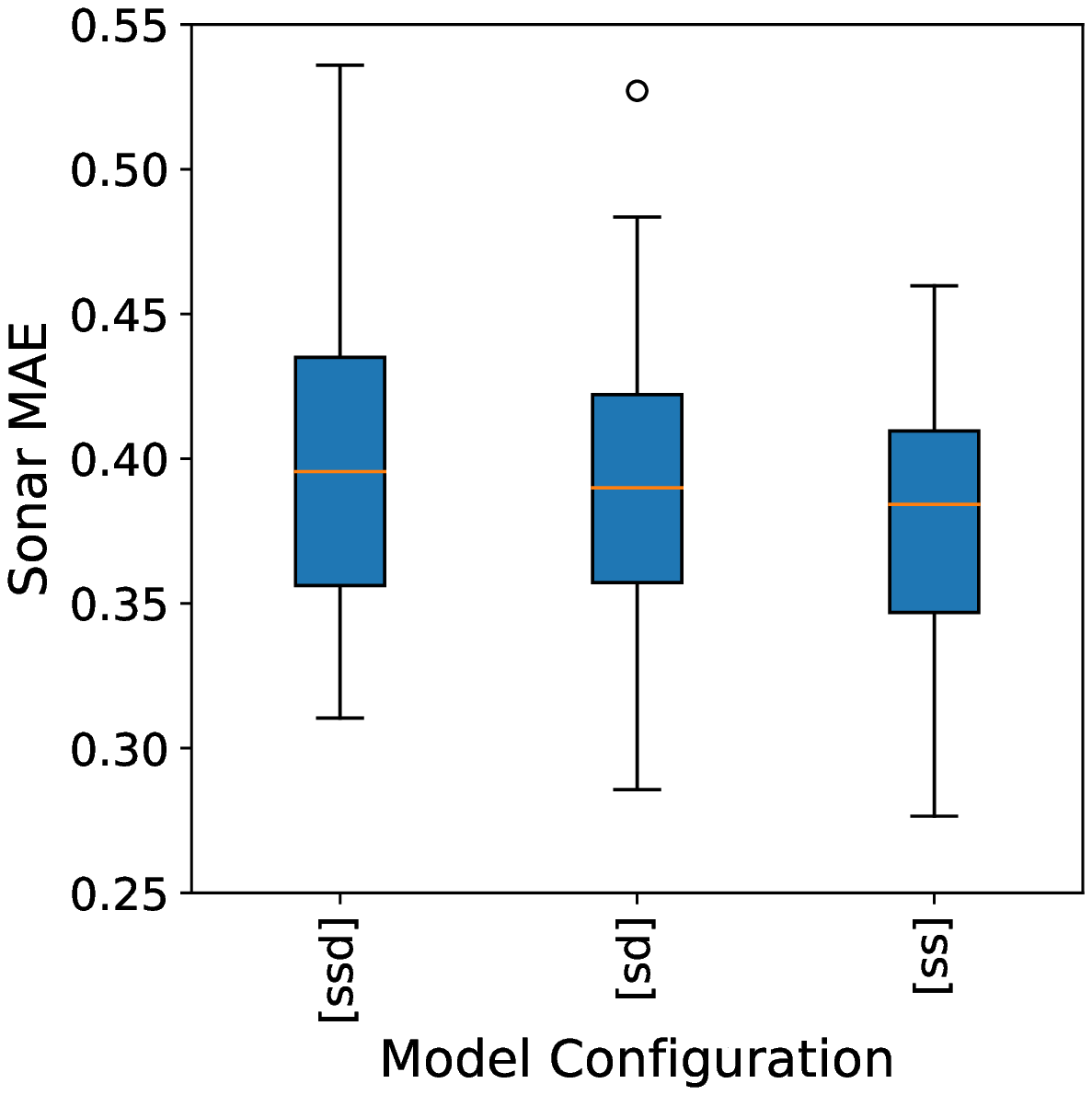}
   \caption{Comparing the LSTM model performance with different feature combinations including discharge for, left) validation dataset, right) test dataset - bridge 742}
  \label{fig: 742-boxplot discharge}
\end{figure}

As shown in Figure \ref{fig: 742-boxplot discharge}, the LSTM performance is not improved by adding discharge as an additional input feature. Replacing stage with discharge did not contribute to better performance either. This outcome confirms the validity of our approach in excluding discharge (or velocity) from the AI models, as it is highly correlated with stage, hence a redundant input feature.

\section{Conclusions}
This study introduces a novel AI approach for the real-time forecast of scour around bridge piers, as a basis for an AI-based early warning system using scour monitoring sensors. Many scour-critical bridges across the US have been monitored since early 2000s using stage and sonar sensors, as well as frequent soundings and surveys. A collaborative research effort with USGS and DOTs enabled access to some of this data. The data from a number of these bridges in Alaska was utilized in this study to train AI algorithms to predict both river stage (flood/water level) and bed elevation variations as early as seven days in advance. The proposed solution was validated using several bridges in Alaska, and case studies were shown for three of these bridges with the most significant scour and flood events. 

LSTMs were trained using historical monitoring data including 11 years of sonar and stage time-series and were shown to provide reliable early forecasts of scour with reasonable variability (uncertainty). Results showed that LSTMs can learn the long-term seasonal patterns in both stage and sonar data and make reasonably accurate predictions for upcoming changes. The uncertainty of predictions was assessed by training the LSTM models for many repetitions. The mean of the predictions was shown to capture the trend of variations of bed elevation during seasonal cycles of scour and filling, happening between July to October each year in Alaska. Also, the upper and lower bound of predictions provided reliable, conservative estimates of scour and filling depth during abrupt changes due to high flows and flood events.

Given that each bridge has its particular trends of scour and flooding depending on the location, bridge structure, river flow, and river bed material and condition, training was performed individually for each bridge. It was shown also through mathematical representation and machine-learning experimentation that LSTMs using the two feature time-series of bed elevation and water level (sonar and stage) can provide scour surrogate models capable of reasonably accurate assessment of scour trend as early as 7 days in advance. The other relevant parameters in typical scour empirical models such as riverbed/soil condition and pier geometry are not temporally variable, therefore excluded from LSTM input features,
allowing for low-dimensional models. The only other relevant feature changing with time, i.e. velocity was shown to be highly correlated with 
stage, also an inferior predictor as compared with stage, hence not included in the final selected input feature combination.

Using GPU clusters, around 300 different configurations of LSTMs were trained and tested to find the optimum algorithm for each bridge. The best configuration showed to vary from one bridge to another, however the optimum LSTM model performance was in the same range with an average scour prediction error between 0.2-0.35m. In comparison with empirical models, the proposed LSTM models offer the following advantages for scour forecast:

\begin{itemize}
\item The LSTM scour forecast models are able to provide real-time prediction of upcoming scour, within a pre-defined forecast window.
\item The average error of scour prediction using the LSTM models is 7.5-25\%, showing a competitive performance against the empirical models with an error range of 7-170\% ~\cite{Sheppard:2014}.
\item The LSTM scour forecast models are developed and trained for individual bridges, therefore inherently aware of the trend of scour particular to a pier location
through learning from historic flow and bed elevation variation patterns. 
\item Unlike the empirical models which involve several parameters, LSTM models are only dependant of two main time-series features, i.e. bed elevation and stage elevation, eliminating the need for measurements of other parameters, such as velocity and particle size.
\end{itemize}

Finally, the cost of implementing the proposed AI-based early warning solution using the already installed stage and sonar sensors for individual bridges is negligible compared to bridge failure costs, especially for critical bridges that are subjected to frequent and/or extreme flooding. It is expected that transportation authorities can benefit from the proposed solution in this study to enhance the existing risk management systems for scour-critical bridges. 
 
\subsection{Data Availability Statement}

Some or all data used for this study are available from the corresponding author upon request. Models, or codes generated or used during the study are proprietary or confidential in nature and may only be provided with restrictions. 

\subsection{Acknowledgments}

This study was funded by Arup Global Research Grant (2020-2021). The authors also would like to thank Tony Marshall, Matt Carter, and Dr Will Cavendish from Arup and Dr Farbod Taymouri from The University of Melbourne, for their valuable advice and support throughout the project.

\bibliography{ref}

\end{document}